\documentclass[a4paper,fleqn]{cas-dc}
\usepackage{graphicx, float}%
\usepackage{multirow}%
\usepackage{amsmath,amssymb,amsfonts}%
\usepackage{amsthm}%
\usepackage{mathrsfs}%
\usepackage[title]{appendix}%
\usepackage{xcolor}%
\usepackage{textcomp}%
\usepackage{manyfoot}%
\usepackage{booktabs}%
\usepackage{algorithm}%
\usepackage{algorithmicx}%
\usepackage{algpseudocode}%
\usepackage{listings}%
\usepackage{array}
\usepackage[position=top]{subcaption}
\usepackage{dcolumn}
\usepackage[utf8]{inputenc} 
\usepackage{soulutf8}
\usepackage{soul}
\usepackage{wrapfig}
\usepackage{soul, color}

\usepackage{makecell}
\usepackage{pifont}
\usepackage{booktabs}

\newcommand{\cmark}{\ding{51}}
\newcommand{\xmark}{\ding{55}}

\usepackage[numbers]{natbib}

\def\tsc#1{\csdef{#1}{\textsc{\lowercase{#1}}\xspace}}
\tsc{WGM}
\tsc{QE}
\tsc{EP}
\tsc{PMS}
\tsc{BEC}
\tsc{DE}
\begin{document}
\let\WriteBookmarks\relax
\def\floatpagepagefraction{1}
\def\textpagefraction{.001}
\shorttitle{Latent Space Guided Scenario Sampling}
\shortauthors{Ulku et~al.}

\title [mode = title]{Latent Space Guided Scenario Sampling for Multimodal Segmentation Under Missing Modalities}                      
\tnotemark[1]

\tnotetext[1]{This document is the results of the research
   project funded by the Scientific and Technological Research Council of Turkey (TUBITAK) under the Grant Number 124E725.}

%\tnotetext[2]{The second %title footnote which is a %longer text matter
%   to fill through the whole %text width and overflow into
%   another line in the %ootnotes area of the first %page.}

\author[1]{Irem Ulku}[type=editor,
                        auid=000,bioid=1,
                        orcid=0000-0003-4998-607X]
\cormark[1]
\ead{irem.ulku@ankara.edu.tr}
\credit{Conceptualization of this study, Methodology, Software}
\affiliation[1]{organization={Department
of Computer Engineering, Ankara University, Ankara,
T{\"u}rkiye}}

\author[1]{{\"O}mer {\"O}zg{\"u}r Tanr{\i}{\"o}ver} [type=editor,
                        auid=000,bioid=1,
                        orcid=0000-0003-0833-3494]
\ead{tanriover@ankara.edu.tr}
\credit{Data curation, Writing - Original draft preparation}

\author[2]{Erdem Akag{\"u}nd{\"u}z}[type=editor,
                        auid=000,bioid=1,
                        orcid=0000-0002-0792-7306]
\ead{akaerdem@metu.edu.tr)}
\credit{Data curation, Writing - Original draft preparation}
\affiliation[2]{organization={Department of Modeling and Simulation, Graduate School of Informatics, METU, Ankara, T{\"u}rkiye}}

\cortext[cor1]{Corresponding author}

%\cortext[cor2]{Principal %corresponding author}

%\fntext[fn1]{This is the %first author footnote, but is %common to third
%  author as well.}

%\fntext[fn2]{Another author %footnote, this is a very long %footnote and
%  it should be a really long %footnote. But this footnote %is not yet
%  sufficiently long enough to %make two lines of footnote %text.}

%\nonumnote{This note has no %numbers. In this work we %demonstrate $a_b$
%  the formation Y\_1 of a new %type of polariton on the %interface
%  between a cuprous oxide %slab and a polystyrene micro-%sphere placed
%  on the slab.}

\begin{abstract}
Multimodal semantic segmentation benefits remote sensing analysis by combining complementary information from different sensor modalities. In real-world remote sensing applications, one or more modalities may be unavailable due to sensor failures, adverse atmospheric conditions, or data acquisition problems. Even with pretrained multimodal representations and existing fine-tuning or adaptation strategies, performance may remain limited because all modality availability scenarios are typically treated as equally informative during training. In this paper, we propose a novel training strategy that learns a scenario sampling distribution directly from the pretrained latent space. Instead of relying on uniform random modality dropout, the proposed method guides fine-tuning toward more informative modality availability scenarios. More specifically, we quantify the effect of each scenario independently based on the distortion it induces in the shared latent representation. We then capture scenario relations using a radial basis function kernel and derive refined scenario scores through a regularized kernel smoothing. These scores are then converted into a probability distribution during scenario sampling for fine-tuning. We evaluate this strategy on three remote sensing image sets, namely DSTL, Potsdam, and Hunan, using CBC-SLP, CBC, and CMX backbones. The experimental results with different image sets and backbones show that our method outperforms standard fine-tuning and LoRA-based adaptation. These findings suggest that the pretrained latent representation can serve as an effective basis for sampling during missing modality fine-tuning. Code is available at https://github.com/iremulku/Latent-Space-Guided-Scenario-Sampling
\end{abstract}

%\includegraphics{figs/cas-%grabs.pdf}


\maketitle

\section{Introduction}

Semantic segmentation in the remote sensing domain aims to assign a semantic category to each pixel in an image, which supports a wide range of Earth observation applications, including land cover mapping \cite{zhang2025flexisam}, environmental monitoring \cite{li2026structfuse}, and urban management \cite{zhou2025emsnet}. Rather than relying on a single modality, multiple heterogeneous modalities acquired by different sensors can be used to exploit complementary cues \cite{ulku2025cross}. For instance, multispectral optical imagery provides discriminative reflectance information, while digital surface model (DSM) or synthetic aperture radar (SAR) data contribute complementary geometric and structural cues that may not be recoverable from optical observations alone \cite{do2025robsense}.

Despite these advantages, in real deployment settings, some modalities may be partially or completely unavailable \cite{han2025multimodal}. Sensor malfunctions, poor atmospheric conditions, and data acquisition failures \cite{das2017deep} may all lead to missing inputs at inference time. As a result, test performance may degrade substantially when one or more modalities are absent \cite{reza2024robust}, especially for a model trained by assuming that all modalities are available. Therefore, many methods have been proposed to address this problem. For instance, MetaRS \cite{zhou2026remote} addresses incomplete remote sensing segmentation by extracting modality-invariant features and using them to recover absent modalities. As a multimodal foundation model for remote sensing, RobSense \cite{do2025robsense} manages incomplete spectral or temporal observations through latent reconstruction. The MHHL framework \cite{han2025multimodal} introduces a heterogeneous hypergraph model to propagate information across samples with incomplete remote sensing modalities. In the context of incomplete remote sensing data fusion, a robust approach \cite{chen2024novel} is proposed to combine modality attention and masked self-attention with contrastive and reconstruction-based pretraining. In general, robustness has also been enhanced by entropy-based regularization \cite{zheng2025reducing}, masked mutual learning \cite{liang2025semantic}, and modality-aware regularization with distillation in MMANet \cite{wei2023mmanet}. However, these studies rely on uniform random modality dropout training policy \cite{zheng2025reducing, liang2025semantic, wei2023mmanet}.

Recently, we developed an alternative architectural design, CBC-SLP \cite{ulku2026robust}, for missing modalities in remote sensing to reduce the performance degradation when all modalities are available at the inference time. Specifically, we structured the latent space into shared and modality-specific components and adaptively transferred them to the decoder according to the modality availability mask. In experiments, we showed that our model outperforms other state-of-the-art approaches across a range of remote sensing image sets with various modality availabilities. However, even with an optimally structured latent space, performance may still be limited because all modality availability scenarios are treated as equally informative during sampling. We hypothesize that during fine-tuning, the modality availability scenarios should be sampled according to how strongly they affect the model’s learned latent space.

Taking inspiration from the MaD-Mix framework \cite{xie2026mad}, we compute scenario descriptors from latent-space behavior, model relations between scenarios through kernelized coupling, and derive a non-uniform training distribution from these scenario effects. Different from MaD-Mix, which reweights training domains for vision-language model training, our strategy reweights modality availability scenarios for multimodal semantic segmentation. In our experiments, this training strategy further improves performance over uniform random modality dropout and surpasses LoRA \cite{hu2022lora} adaptation baselines. These results indicate that directly exploiting the pretrained latent structure provides more informative and task-relevant guidance for training than generic parameter-efficient fine-tuning \cite{ma2025unified, ma2025sasam}.

The main contributions of this paper are summarized as follows:
\begin{itemize}
    \item We establish a novel training strategy for multimodal semantic segmentation when some modalities are missing. It derives scenario weights directly from the learned latent space and replaces uniform random modality dropout with a learned distribution over modality availability scenarios.

     \item We define scenario importance through shared latent distortion and convert these raw scenario statistics into a sampling distribution by solving a regularized kernel smoothing problem over the scenario space. This formulation captures the relative effect of each scenario and also the relations between scenarios in the latent descriptor space.

    \item We evaluate the proposed strategy on the DSTL, Potsdam, and Hunan remote sensing image sets using CBC-SLP, CBC, and CMX, and compare it against standard training with random modality dropout and LoRA-based adaptation. The results show that the proposed strategy provides more consistent improvements across image sets and models.
\end{itemize}

\section{Related Works}

\subsection{Multimodal Semantic Segmentation with Missing Modalities}

Semantic segmentation under missing modalities has gained significant popularity because incomplete inputs remain a problem for multimodal systems  in real-world applications. Existing studies have addressed this problem from several perspectives. Some methods recover absent information through reconstruction, synthesis, or modality completion \cite{zhou2026remote, do2025robsense}. Others improve robustness to incomplete information by learning modality-invariant or shared representations through explicit cross-modal correlation modeling \cite{zhang2022mmformer, qiu2024mmmvit, chen2024novel, han2025multimodal, wei2023mmanet}. Meanwhile, several methods strengthen robustness through modified training objectives, including distillation, entropy-based regularization, mutual learning, and modality-aware regularization \cite{zheng2025reducing, liang2025semantic, wei2023mmanet}.

For instance, the MMANet \cite{wei2023mmanet} framework combines margin-aware distillation and modality-aware regularization strategies to assist incomplete multimodal learning. To balance the contribution of each modality, \cite{zheng2025reducing} introduces a regularization term based on functional entropy. A masked mutual learning scheme has been proposed in \cite{liang2025semantic} to ensure multi-level consistency and enhance mutual learning in incomplete multimodal scenarios. The mmFormer architecture \cite{zhang2022mmformer} learns modality-invariant features by combining modality-specific encoders with an inter-modal Transformer that builds and aligns long-range relations across modalities. MMMViT \cite{qiu2024mmmvit} improves incomplete modality segmentation performance by explicitly leveraging modality correlations before shared-representation fusion. These studies demonstrate that robustness under missing modalities can be improved through more effective representation design and training objectives. However, the sampling distribution in the training phase still tends to treat modality availabilities as equally important, potentially overlooking how differently these scenarios affect the learned latent representation.

\subsection{Multimodal Remote Sensing Segmentation with Missing Modalities}

Although multimodal semantic segmentation with missing or incomplete modalities has been extensively studied in the medical imaging domain \cite{zhang2022mmformer, qiu2024mmmvit,  shi2026addressing, zhang2026disentangling}, its potential in remote sensing remains underexplored. In one relevant study \cite{zhou2026remote}, MetaRS  addresses incomplete remote sensing segmentation by extracting modality-invariant meta-modal features and using them to recover absent modalities while regularizing the learned representation for the target task. As a robust multimodal foundation model, RobSense \cite{do2025robsense} manages incomplete spectral or temporal inputs through unimodal latent reconstructors trained with
masked spectral bands and time points. The MHHL \cite{han2025multimodal} framework enables incomplete modalities to learn features from complete modalities by incorporating heterogeneous hypernodes that capture relationships among different modality combinations. In addition, a robust remote sensing fusion approach \cite{chen2024novel} combines modality attention and masked self-attention mechanisms, and employs reconstruction and contrastive loss to maintain high performance when dealing with incomplete inputs.

These studies demonstrate the advantage of dedicated representation learning or training mechanisms for remote sensing segmentation with missing modalities rather than traditional fusion methods \cite{li2025semantic, zhang2023cmx, ulku2025cross}. More recently, our CBC-SLP \cite{ulku2026robust} extends this line by structuring the latent space into shared and modality-specific components and adaptively routing private features according to the modality available scenarios. While CBC-SLP achieves state-of-the-art test set performance across full and missing modalities, it still relies on uniform random modality dropout during training. Consequently, a critical challenge is to develop principled techniques that learn the scenario distribution from the pretrained latent structure rather than treating all modality scenarios as equally important \cite{dong2026advances}.

\subsection{Training Strategies for Multimodal Segmentation with Incomplete Modalities}

Instead of addressing incomplete multimodal segmentation only through architecture design, recent studies have increasingly explored how training itself should be adapted. This has led to a new line of methods that improve robustness through adaptive optimization and modality-aware training strategies. To jointly address unimodal and multimodal imbalance, hierarchical gradient alignment \cite{shi2026addressing} is proposed as a meta-learning-inspired optimization strategy. Their method enables more stable learning for missing modalities. MECS-Net \cite{chen2026towards} employs a task-oriented sample-level modal contribution evaluation and resampling optimization strategy to guide the training process towards high-contribution modality combinations. Additionally, it enhances the representational capacity of low-contribution modalities by a sample-level resampling. CHARM \cite{wen2026charm} introduces a complementary learning framework for discovering modality-interactive correspondences, while preserving modality-specific characteristics through a dual-path optimization strategy. DANTE \cite{zhang2026disentangling} introduces a two-stage framework, first adopting a self-supervised pretraining to improve feature robustness for absent modalities, followed by information-theoretic regularization to maximize the transfer of rich information. To integrate data from multiple modalities in the remote sensing domain when modality-aligned data is limited, two novel self-supervised learning approaches are proposed in \cite{linial2025enhancing}. By leveraging a mixing strategy during pretraining, this study mitigates spatial misalignment errors and improves model robustness. In a related line of research, a parameter-efficient adaptation procedure \cite{reza2024robust} for multimodal learning modulates intermediate features using scaling, shifting, or low-rank transformations to improve robustness to missing modalities. Overall, these studies demonstrate that robustness to missing modalities can be enhanced through specifically designed training strategies.

Recently, MaD-Mix \cite{xie2026mad} showed that multimodal vision-language model training need not rely on manually specified data mixtures, but can instead be guided by a specialized latent-space formulation that assigns sampling weights to different training domains. To address incomplete data, they propose explicitly decoupling missing modalities from the optimization process, ensuring no associated noise is propagated into the alignment objective. Their results are obtained by casting multimodal data mixing as a latent-space coupling objective and deriving closed-form alignment scores through a dual solution. An important question arises in this context: If a shared latent space can reveal the relative importance of multimodal training domains, can it also be used to determine the relative importance of modality scenarios for incomplete multimodal segmentation? Our research addresses this question by studying how scenario importance can be derived directly from the learned latent space of a pretrained segmentation model. We examine whether the latent structure can provide a more informative basis for training than uniform random modality dropout, and use it to construct a scenario-aware training distribution for multimodal segmentation with incomplete modalities.

%+The rest of the paper is organized as follows, Section 2... \hl{......}

\section{Methodology}

\begin{figure*}[t]
\centering
\includegraphics[width=\textwidth]{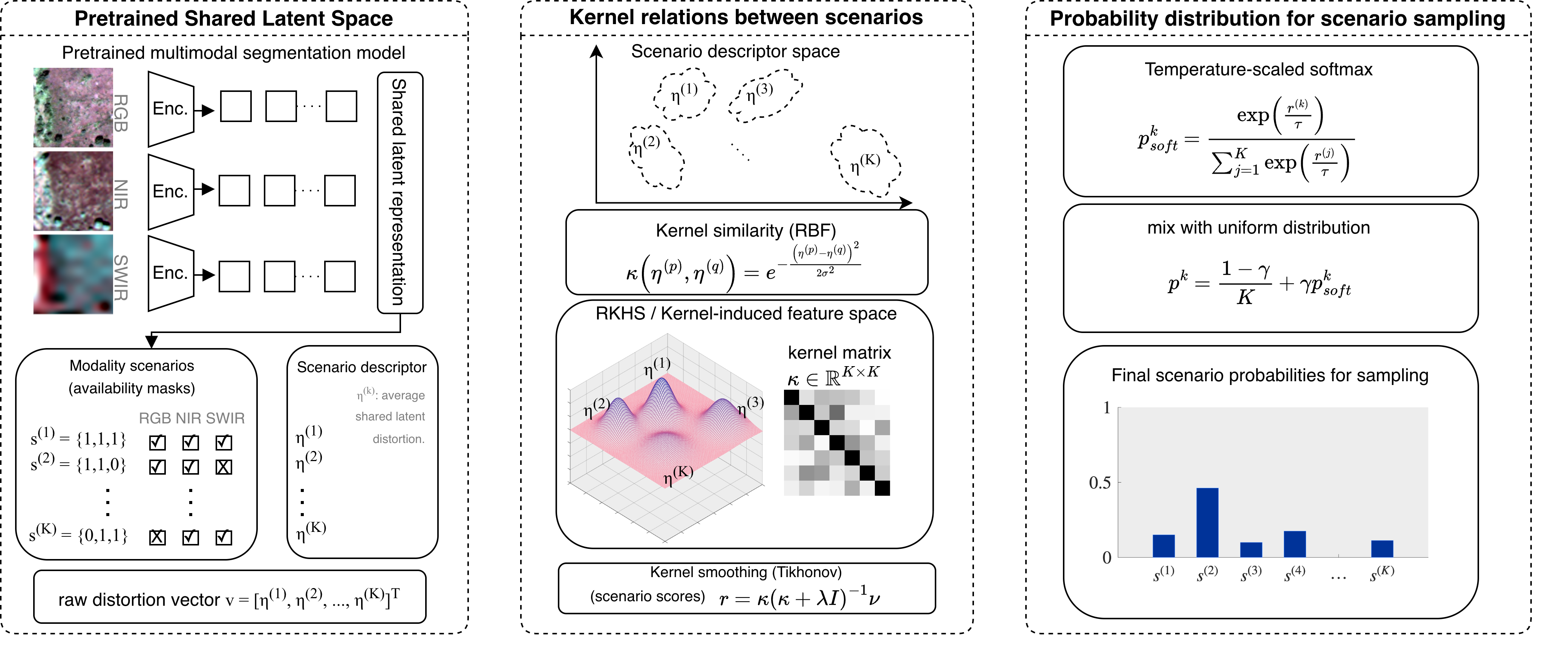}
\caption{Overview of the latent-space-guided scenario sampling framework.}
\label{process}
\end{figure*}

We propose a fine-tuning strategy for multimodal semantic segmentation in the presence of missing modalities. An overview of our strategy is shown in Fig.~\ref{process}. The main idea is to replace the uniform scenario sampling used in standard random modality dropout with a probability distribution over modality availability scenarios. The proposed distribution is obtained from the latent behavior of a pretrained multimodal segmentation model and is then used during the fine-tuning stage.

The proposed strategy is inspired by the MaD-Mix framework ~\cite{xie2026mad}, which derives sampling weights through a regularized kernel operator defined on multimodal representations. However, unlike MaD-Mix, which computes sampling weights over multimodal data domains, our strategy computes sampling probabilities over modality availability scenarios. Moreover, since a scenario is a masking condition applied to the same multimodal data rather than a standalone semantic domain, scenario importance is defined through the distortion induced in the shared latent space of the pretrained segmentation model.

The proposed strategy will be described on the CBC-SLP architecture shown in Fig.~\ref{model}. Although the derivation in this section is presented for the CBC-SLP model, the same training strategy is not restricted to this architecture. For models that only use a shared latent representation, such as the original CBC model, the formulation is directly applicable without modification.

\subsection{Problem Definition}

The input sample is denoted by $\mathbf{X} = [\mathbf{X}_1, \mathbf{X}_2, \ldots, \mathbf{X}_M]$, where $M$ is the number of modalities and $\mathbf{X}_m$ is the input tensor of modality $m$. Here, $\mathcal{M}$ denotes the set of all modalities as $\mathcal{M} = \{1,2,\ldots,M\}$. To model missing modalities, a binary modality availability mask is introduced as
\begin{equation}
\mathbf{s} \in \{0,1\}^{B \times M},
\label{eq:mask}
\end{equation}
where $B$ denotes the batch size, $s_{b,m}=1$ indicates that modality $m$ is available for sample $b$, and $s_{b,m}=0$ indicates that modality $m$ is missing.

In standard random modality dropout, modality availability scenarios are sampled uniformly. However, the proposed strategy assigns a non-uniform probability to each scenario. The probability distribution is determined by how strongly a scenario disrupts the latent representation of the pretrained model.

\subsection{Structured Latent Representation}

\begin{figure*}[t]
\centering
\includegraphics[width=\textwidth]{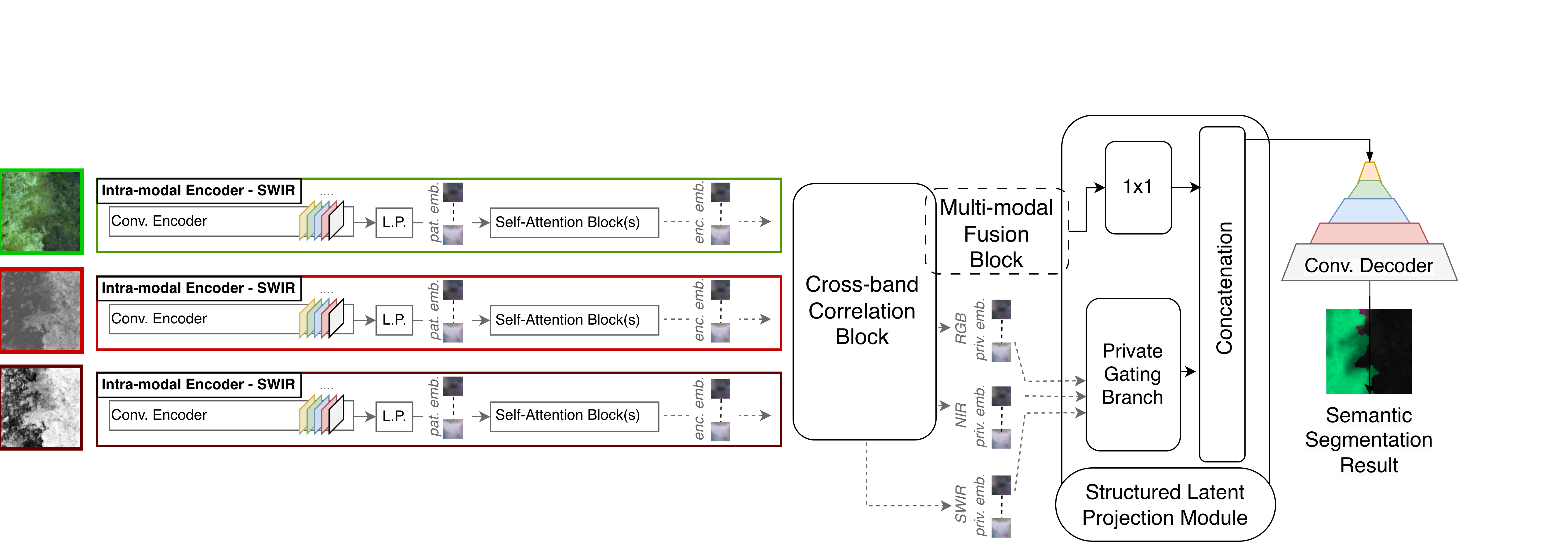}
\caption{Overview of CBC-SLP pipeline.}
\label{model}
\end{figure*}

The CBC-SLP architecture \cite{ulku2026robust} shown in Fig.~\ref{model} includes both shared and modality-specific latent branches. In the proposed training strategy, scenario weighting is computed from the shared latent representation. Let $\mathbf{X}_6^{inter}$ denote the deep inter-modal fused latent representation. Then, the shared latent representation is defined as
\begin{equation}
\mathbf{z}^{sh} = \mathrm{Conv}^{1\times1\times1}_{sh}\!\left(\mathbf{X}_6^{inter}\right),
\label{eq:shared}
\end{equation}
where $\mathrm{Conv}^{1\times1\times1}_{sh}(\cdot)$ denotes a learnable $1\times1\times1$ projection layer. This shared latent tensor provides a compact representation of the multimodal fused information before decoding and serves as the basis for the proposed scenario scoring mechanism.

\subsection{Missing Modality Scenarios}

The set of modality availability scenarios can be defined as
\begin{equation}
\mathcal{S} = \left\{\mathbf{s}^{(k)}\right\}_{k=1}^{K}, \qquad \mathbf{s}^{(k)} \in \{0,1\}^{M},
\label{eq:scenario_set}
\end{equation}
where $K = 2^{M}-1$ is the number of modality combinations and $\mathbf{s}^{(k)}$ denotes the $k^{th}$ modality availability scenario.

In conventional random modality dropout, each scenario in $\mathcal{S}$ is sampled uniformly. The limitation of this strategy is that all scenarios are treated as equally informative, although in practice, some missing modality patterns disrupt the latent representation more than others.

\subsection{Full Modality Reference State}

To quantify the effect of each scenario, we use the shared latent representation obtained under full modality availability as a reference. Let $\mathbf{z}^{sh}(\mathbf{X},\mathbf{s})$ denote the shared latent representation produced by the model for input $\mathbf{X}$ under modality mask $\mathbf{s}$. If $\mathbf{F}_M = [1,1,\ldots,1]\in \{1\}^{M}$ denotes the full modality mask, then the forward pass for a training sample $\mathbf{X}_i$ yields
\begin{equation}
\mathbf{z}_{i}^{sh,full} = \mathbf{z}^{sh}(\mathbf{X}_i,\mathbf{F}_M),
\label{eq:full_shared}
\end{equation}
where $\mathbf{z}_{i}^{sh,full}$ is the shared latent representation of sample $i$ under full modality availability. Since the full modality case preserves the complete shared multimodal information, deviations from this reference indicate how much the shared latent representation is altered under a given missing modality scenario.

\subsection{Shared Latent Distortion}

This scenario statistic measures how much a scenario disrupts the shared latent representation. For sample $i$ and scenario $\mathbf{s}^{(k)}$, the shared latent distortion is defined as
\begin{equation}
d_{sh}^{(k)}(i) =
\frac{1}{|\Omega_{sh}|}
\left\|
\mathbf{z}^{sh}(\mathbf{X}_i,\mathbf{s}^{(k)}) - \mathbf{z}^{sh}(\mathbf{X}_i,\mathbf{F}_M)
\right\|_2^2,
\label{eq:dsh}
\end{equation}
where $|\Omega_{sh}|$ denotes the total number of elements in the shared latent tensor and $\|\cdot\|_2^2$ expresses the squared Euclidean norm over all entries.

Equation~(\ref{eq:dsh}) measures the mean-squared deviation of the shared latent tensor under scenario $\mathbf{s}^{(k)}$ from the full modality shared latent tensor. A larger value indicates that the scenario causes a stronger degradation in the shared representation.

The average shared distortion of scenario $k$ over the training set is then given by
\begin{equation}
D_{sh}^{(k)} = \frac{1}{N}\sum_{i=1}^{N} d_{sh}^{(k)}(i),
\label{eq:Dsh}
\end{equation}
where $N$ is the number of samples used to estimate the scenario statistics.

\subsection{Scenario Importance Score}

Since the proposed formulation is based only on the shared latent branch, the importance of scenario $k$ is defined directly from the average shared latent distortion as
\begin{equation}
\eta^{(k)} = D_{sh}^{(k)}.
\label{eq:etak}
\end{equation}

A larger value of $\eta^{(k)}$ indicates that scenario $k$ induces a stronger distortion in the shared latent representation. Thus, $\eta^{(k)}$ measures the severity of scenario $k$ from the perspective of the pretrained model.

\subsection{Kernelized Scenario Coupling}

The proposed strategy is inspired by the MaD-Mix framework ~\cite{xie2026mad}, which computes sampling weights through a regularized kernel operator defined on multimodal domain embeddings. In our strategy, this idea is adapted to modality availability scenario reweighting. Specifically, we assign scores to missing modality scenarios based on their latent effect on the pretrained segmentation model.

To capture similarity relations between scenarios, we define a radial basis function (RBF) kernel ~\cite{scholkopf1997comparing} over the scenario descriptors. That is, beyond the individual raw distortions, we also model the relations between scenarios in the descriptor space. We choose the RBF kernel because our goal is to assign similar refined scores to scenarios that induce similar levels of shared latent distortion. The RBF kernel provides such a distance-based coupling by assigning high similarity to nearby descriptors and gradually decreasing similarity as their latent effects diverge.  In addition, since the RBF kernel corresponds to an inner product in an infinite-dimensional space, it can capture nonlinear relations among scenario descriptors ~\cite{momeni2025continual}. Accordingly, the RBF kernel over the scenario descriptors is defined as follows:
\begin{equation}
\kappa\!\left(\eta^{(p)}, \eta^{(q)}\right)
=
\exp\left(
-\frac{\left(\eta^{(p)} - \eta^{(q)}\right)^2}{2\sigma^2}
\right),
\label{eq:kernel}
\end{equation}
\begin{equation}
p,q \in \{1,\ldots,K\},
\nonumber
\end{equation}
where $\sigma > 0$ is the kernel bandwidth parameter. Thus, two scenarios receive a high kernel value when they induce similar levels of shared latent distortion.

Let $\mathcal{K} = [\mathcal{K}_{pq}] \in \mathbb{R}^{K \times K}$ denote the scenario kernel matrix, where
\begin{equation}
\mathcal{K}_{pq}
=
\kappa\!\left(\eta^{(p)}, \eta^{(q)}\right),
\label{eq:kernel_matrix_entry}
\end{equation}
and let the corresponding raw scenario importance vector be expressed as follows:
\begin{equation}
\boldsymbol{\nu} =
\begin{bmatrix}
\eta^{(1)} & \eta^{(2)} & \cdots & \eta^{(K)}
\end{bmatrix}^{\top}
\in \mathbb{R}^{K}.
\label{eq:nuvec}
\end{equation}

The kernel matrix $\mathcal{K}$ encodes pairwise relations between scenarios, while the vector $\boldsymbol{\nu}$ collects their raw importance values. We use these two quantities to estimate a refined scoring function over the scenario space, whose evaluations at the scenario descriptors define the final scenario score vector.

Therefore, the final scenario score vector can be obtained by solving a regularized kernel smoothing problem over the scenario space ~\cite{gong2024supervised}. We seek a refined scoring function that remains close to these raw values while varying smoothly across related scenarios. For this purpose, the scenario descriptors serve as inputs to a kernel-induced function space, which allows the scoring function to be defined in terms of the similarities encoded by the kernel. In particular, we define $\mathcal{H}_{\kappa}$ as the reproducing kernel Hilbert space (RKHS) \cite{rosipal2001kernel} spanned by the kernel in equation (\ref{eq:kernel}). RKHS is a Hilbert space of functions associated with a positive semidefinite kernel. Its key advantage in our study is that the scoring function can be constructed from kernel similarity values between scenario descriptors. This property enables the estimation of a smooth scoring function directly from pairwise similarities between scenarios.

Hence, we denote $f \in \mathcal{H}_{\kappa}$ as a scoring function defined on the scenario descriptor space, where $f(\eta^{(k)})$ represents the refined score assigned to scenario $k$. Formulating the RKHS regression problem in ~\cite{tsai2023sample}, we define the scoring function in the RKHS associated with the chosen kernel. This allows the function to be estimated directly from the kernel similarities among the observed scenario descriptors. To obtain a function that both fits the raw scenario importance values and remains smooth over the scenario space, we consider the following regularized objective:
\begin{equation}
\min_{f \in \mathcal{H}_{\kappa}}
\frac{1}{2}\|f\|_{\mathcal{H}_{\kappa}}^{2}
+
\frac{1}{2\lambda}\sum_{k=1}^{K}\left(\eta^{(k)} - f(\eta^{(k)})\right)^{2},
\label{eq:rkhs_obj}
\end{equation}
where $\|f\|_{\mathcal{H}_{\kappa}}^2$ denotes the RKHS norm of $f$, $\mathcal{H}_{\kappa}$ is the reproducing kernel Hilbert space defined by the kernel in (\ref{eq:kernel}), and $\lambda > 0$ is a regularization coefficient. The first term in (\ref{eq:rkhs_obj}) penalizes overly complex or irregular score functions, so that the resulting score function varies smoothly over the scenario space. The second term keeps the estimated scores close to the raw scenario importance values $\{\eta^{(k)}\}_{k=1}^{K}$.

By the representer theorem \cite{dinuzzo2012representer}, for regularized objectives of the form in equation (\ref{eq:rkhs_obj}), whose dependence on $f$ is restricted to the RKHS norm and the function values at the observed scenario descriptors, the minimizer does not need to be sought over the entire infinite-dimensional RKHS. Instead, the minimizer lies in the finite-dimensional subspace spanned by the kernel evaluations at the observed scenario descriptors, and therefore can be expressed with the following finite kernel expansion:
\begin{equation}
f(\cdot)=\sum_{j=1}^{K} a_j \kappa(\eta^{(j)},\cdot),
\label{eq:representer}
\end{equation}
where $\mathbf{a}=[a_1,\ldots,a_K]^\top$ denotes expansion coefficients. Evaluating the representer expansion in equation (\ref{eq:representer}) at the scenario descriptors yields the score vector as $\mathbf{r}=\mathcal{K}\mathbf{a}$. Furthermore, under this expansion, the RKHS norm becomes $\|f\|_{\mathcal{H}_{\kappa}}^2=\mathbf{a}^\top \mathcal{K}\mathbf{a}$. This follows from the inner-product structure of finite RKHS expansions.  Substituting these expressions into equation (\ref{eq:rkhs_obj}) reduces the problem to a finite-dimensional quadratic optimization over the coefficient vector $\mathbf{a}$. Solving the corresponding first-order optimality condition gives $\mathbf{a}=(\mathcal{K}+\lambda\mathbf{I})^{-1}\boldsymbol{\nu}$, and therefore the kernel-smoothed scenario score vector becomes as follows:
\begin{equation}
\mathbf{r} =
\mathcal{K}\left(\mathcal{K} + \lambda \mathbf{I}\right)^{-1}\boldsymbol{\nu},
\label{eq:rvec}
\end{equation}
where $\mathbf{I}$ is the $K \times K$ identity matrix and $\mathbf{r}\in\mathbb{R}^{K}$ contains the final kernel-smoothed scores of all scenarios.

Equation~(\ref{eq:rvec}) ensures that the proposed method does not score each scenario independently. Instead, the raw severity of a scenario is smoothed according to its similarity to other scenarios in the shared latent descriptor space. Therefore, scenarios with similar latent disturbance levels receive similar scores.

\subsubsection*{Spectral Interpretation}

Since $\mathcal{K}$ is a symmetric kernel matrix, the spectral theorem for symmetric matrices implies that it is orthogonally diagonalizable \cite{trouillon2017knowledge}. Therefore, $\mathcal{K}$ can be written as
\begin{equation}
\mathcal{K} = \mathbf{U}\mathbf{\Sigma}\mathbf{U}^{\top},
\label{eq:eig}
\end{equation}
where $\mathbf{\Sigma}=\mathrm{diag}(\sigma_1,\ldots,\sigma_K)$ contains the eigenvalues of $\mathcal{K}$ and the columns of $\mathbf{U}$ are the corresponding orthonormal eigenvectors. Substituting (\ref{eq:eig}) into (\ref{eq:rvec}) results the following:
\begin{equation}
\mathbf{r}
=
\mathbf{U}
\mathrm{diag}\left(
\frac{\sigma_1}{\sigma_1+\lambda},
\ldots,
\frac{\sigma_K}{\sigma_K+\lambda}
\right)
\mathbf{U}^{\top}\boldsymbol{\nu}.
\label{eq:spectral_r}
\end{equation}

Equation~(\ref{eq:spectral_r}) shows that the proposed kernel operator applies a spectral soft-thresholding filter to the raw scenario importance vector. Directions associated with large eigenvalues are preserved more strongly, whereas directions associated with small eigenvalues are attenuated. Hence, the final scenario scores emphasize the main directions of variation in the scenario space while suppressing noisy directions.

The advantage of the proposed approach over direct heuristic weighting is that it accounts not only for the magnitude of scenario importance, but also for the relations between scenarios in the latent descriptor space. As a result, similar scenarios are assigned more consistent scores, which reduces sensitivity to noisy estimates and yields a smoother probability distribution.

\subsection{Probability Distribution over Scenarios}

The scenario scores are converted into probabilities in three steps. First, the score vector is standardized as follows:
\begin{equation}
\bar r^{(k)} =
\frac{r^{(k)} - \mu_r}{\sigma_r},
\qquad k = 1,\ldots,K,
\label{eq:rbar}
\end{equation}
where
\begin{equation}
\mu_r = \frac{1}{K}\sum_{k=1}^{K} r^{(k)},
\qquad
\sigma_r =
\sqrt{
\frac{1}{K}\sum_{k=1}^{K}\left(r^{(k)}-\mu_r\right)^2
}.
\label{eq:rstats}
\end{equation}
Here, $\mu_r$ and $\sigma_r$ denote the mean and standard deviation of the scenario scores, respectively.

Second, a temperature-scaled softmax is applied to convert the scenario scores into probabilities as follows:
\begin{equation}
p_{soft}^{(k)} =
\frac{\exp\left(\tau \bar r^{(k)}\right)}
{\sum_{j=1}^{K}\exp\left(\tau \bar r^{(j)}\right)},
\qquad k = 1,\ldots,K,
\label{eq:psoft}
\end{equation}
where $\tau > 0$ is the temperature parameter. This scaling is used to avoid sharp probability assignments, resulting in a smoother distribution across scenarios.

Third, a convex combination of the softmax distribution and the uniform distribution is used to incorporate the obtained scenario preference while maintaining a balanced sampling pattern:
\begin{equation}
p^{(k)} =
(1-\gamma)\frac{1}{K}
+
\gamma p_{soft}^{(k)},
\qquad k = 1,\ldots,K,
\label{eq:pk}
\end{equation}
where $\gamma \in [0,1]$ is a mixing coefficient controlling the relative contribution of the uniform distribution and the softmax distribution. The uniform component in equation (\ref{eq:pk}) ensures that all scenarios remain observable during fine-tuning and prevents the training from collapsing to only a small subset of modality availability patterns. Therefore, the probabilities are normalized to ensure a valid probability distribution as
\begin{equation}
\sum_{k=1}^{K} p^{(k)} = 1.
\label{eq:psum}
\end{equation}

\subsection{Probability Guided Training Loss Function}

Let $\ell_{seg}(\mathbf{X},\mathbf{Y};\mathbf{s},\theta)$ denote the segmentation loss of the model with parameters $\theta$ for input $\mathbf{X}$ and ground-truth mask $\mathbf{Y}$ under scenario mask $\mathbf{s}$. By using standard random modality dropout, the expected training loss is defined as follows:
\begin{equation}
\mathcal{L}_{uni}(\theta) =
\mathbb{E}_{(\mathbf{X},\mathbf{Y})\sim\mathcal{D},\;\mathbf{s}\sim\mathrm{Unif}(\mathcal{S})}
\left[
\ell_{seg}(\mathbf{X},\mathbf{Y};\mathbf{s},\theta)
\right],
\label{eq:Luni}
\end{equation}
where $\mathcal{D}$ denotes the training distribution and $\mathrm{Unif}(\mathcal{S})$ denotes uniform sampling over all valid scenarios.

In contrast, the proposed training loss uses the scenario probabilities derived from the latent descriptor space as follows:
\begin{equation}
\mathcal{L}_{prob}(\theta) =
\mathbb{E}_{(\mathbf{X},\mathbf{Y})\sim\mathcal{D},\;\mathbf{s}\sim p}
\left[
\ell_{seg}(\mathbf{X},\mathbf{Y};\mathbf{s},\theta)
\right],
\label{eq:Lprob}
\end{equation}
where $p$ denotes the non-uniform scenario distribution obtained from (\ref{eq:pk}).

Equation~(\ref{eq:Lprob}) means that the model is fine-tuned more often on scenarios that induce stronger latent distortions, while still retaining exposure to the remaining scenarios through the uniform component in equation (\ref{eq:pk}).

The method includes in two stages. First, the multimodal segmentation model is trained with the random modality dropout strategy. Finally, the model is fine-tuned for a small number of additional epochs by sampling scenario masks according to the learned distribution $p$ instead of the uniform distribution.

\subsection{Discussion}

Fig.~\ref{process} illustrates the overall latent-space-guided scenario sampling framework. The motivation is that different modality availability scenarios disrupt the shared latent branch to different degrees. Uniform random modality dropout ignores this heterogeneity. The proposed method instead estimates scenario importance directly from the shared latent behavior of the pretrained model and then uses a kernelized probability distribution to bias fine-tuning toward the most informative scenarios.

MaD-Mix framework ~\cite{xie2026mad} derives its regularized kernel operator from a modality-aware primal-dual alignment objective over data domains. In contrast, the proposed method uses a regularized kernel smoothing formulation over modality availability scenarios. Since a scenario is not itself a semantic data domain but rather a masking condition applied to the same data, the proposed formulation measures how strongly each scenario disrupts the pretrained shared latent representation instead of defining scenario importance through domain-level alignment.

The proposed strategy is a general training mechanism for multimodal segmentation models trained with random modality dropout. Since it relies only on the shared latent behavior of a pretrained model, it can be applied to multimodal models that operate entirely on a shared latent representation.

\section{Experiments}

We evaluate the performance of the proposed sampling strategy on three multimodal remote sensing image sets. Our goal is to determine whether replacing uniform scenario sampling with a latent-space-guided distribution during fine-tuning yields better segmentation results across all possible modality availability scenarios. We consider three prevalent models on missing multimodal semantic segmentation, namely CBC-SLP \cite{ulku2026robust}, CBC \cite{ulku2025cross}, and CMX \cite{zhang2023cmx}. 

The pretrained models are fine-tuned using four settings. \textbf{No fine-tuning} is the baseline, where the models are trained for 70 epochs with uniform random modality dropout. In this setting, we directly report the results from our previous experiments \cite{ulku2026robust} without fine-tuning. \textbf{Fine-tune with random modality dropout} represents fine-tuning the pretrained baseline models for 10 epochs using a random modality dropout sampling policy. \textbf{Fine-tune with LoRA-based adaptation} provides a parameter-efficient fine-tuning method applied to pretrained baseline models for 10 epochs with the same uniform sampling. In this setting, LoRA updates are restricted to selected deeper multimodal fusion layers, whereas the remaining pretrained parameters are kept fixed. Lastly, the \textbf{Fine-tune with our strategy} replaces uniform scenario sampling during 10 epochs of fine-tuning with the probability distribution estimated from the pretrained latent space. Each experiment is conducted as a binary semantic segmentation task.

\subsection{Image sets}

To evaluate the proposed training strategy under both homogeneous and heterogeneous modalities, we conduct experiments on three public remote sensing image sets. These sets contain information from different sensor types, spatial resolutions, and modality compositions. For each image set, the target class is selected based on sufficient annotation and high responsiveness to the modality-specific information from the available inputs.

\paragraph{DSTL}
The DSTL \cite{dstl_kaggle} image set consists of 25 image tiles from the WorldView-3 optical Earth observation satellite, each covering approximately 1 km $\times$ 1 km. We chose this image set because it provides homogeneous optical images as RGB, near-infrared (NIR), and shortwave infrared (SWIR), allowing us to examine the performance of our strategy under spectral modalities. The visible RGB bands cover the 450-690 nm wavelength region, the NIR bands span 770-1040 nm, and the SWIR modality consists of eight bands in the 1195-2365 nm range. These modalities are acquired at different spatial resolutions, namely about 0.31 m for the visible bands, 1.24 m for the multispectral bands, and approximately 7.5 m for SWIR. The crop category is selected as the target class because it is sufficiently annotated and benefits strongly from the discriminative information in the NIR and SWIR bands.

\paragraph{Potsdam}
The Potsdam \cite{isprs_potsdam} is a high-resolution aerial image set for semantic segmentation provided by the International Society for Photogrammetry and Remote Sensing. It consists of 38 images, each of size 6000 $\times$ 6000 pixels, with a ground sampling distance of 0.05 m. In our experiments, Potsdam is used as a heterogeneous trimodal image set composed of RGB, IRRG (NIR-R-G), and DSM modalities. The tree category typically provides sufficient annotation and is highly responsive to the complementary modalities, making it a suitable target class for evaluation.

\paragraph{Hunan}
The Hunan \cite{li2022dkdfn} is a heterogeneous trimodal image set collected over Hunan Province, China. It combines Sentinel-2 multispectral optical imagery, Sentinel-1 SAR observations, and topographic data, which are used in this work as MSI, SAR, and DEM modalities, respectively. Sentinel-2 provides spectral observations in the 492.7-2202.4 nm wavelength range, covering visible, NIR, and SWIR bands, while Sentinel-1 contributes dual-polarization SAR measurements. The topographic modality contains both elevation and slope sources from the Shuttle Radar Topography Mission (SRTM). The spatial resolutions differ across modalities, such that 10 m and 20 m for Sentinel-2, 10 m for Sentinel-1, and 30 m for the DEM. Hunan has 500 images covering 32,768,000 pixels. Due to its extensive spatial coverage and high sensitivity to the provided modalities, the forest category is selected as the target class.

\subsection{Training settings}

As part of the preprocessing pipeline, all images are divided into $224 \times 224$ patches. Input normalization is performed separately for each modality by computing the per-channel mean only from the training subset and applying the resulting statistics to the training, validation, and test samples. For DSTL and Potsdam, the data is split into 72\% training, 8\% validation, and 20\% test sets. For Hunan, the official split scheme of 80\% training, 10\% for validation, and 10\% for testing is used.

The implementation is based on PyTorch, and all experiments are conducted on the NVIDIA RTX 6000 Ada GPU. For each baseline model, the training phase lasts 70 epochs and uses the Adam optimizer with an initial learning rate of $1 \times 10^{-4}$, a mini-batch size of 8, and binary cross-entropy loss. The learning rate is reduced by 9\% every 5 epochs throughout training. To reduce sensitivity to a particular data partition, a 5-fold cross-validation protocol is used in all experiments.

After the initial training stage, each pretrained model is further trained for 10 additional epochs under one of the three fine-tuning settings explained above. In the \textbf{Fine-tune with random modality dropout} setting, the same uniform random scenario sampling used in the original training stage is retained. In the \textbf{Fine-tune with LoRA-based adaptation} setting, the pretrained weights are frozen, and only low-rank residual adapters inserted into selected deeper multimodal fusion layers are updated. In the \textbf{Fine-tune with our strategy} setting, the fine-tuning is again fixed to 10 epochs, but the modality availability scenarios are drawn from the probability distribution estimated from the pretrained latent space. Therefore, all comparisons share the same pretrained backbone and the same number of epochs to directly measure the contribution of the proposed fine-tuning strategy on the final performance. 

For scenario probability estimation, the RBF kernel bandwidth and the regularization coefficient are set to $\sigma=1.0$ and $\lambda=10^{-3}$, respectively. The temperature-scaled softmax uses $\tau=0.5$, and the final distribution is mixed with the uniform distribution using $\gamma=0.5$. Because all image sets contain three modalities, probability estimation is performed over the $2^3-1=7$ valid scenarios for modality availability. The scenario statistics are computed on the training split with a batch size of 1.

For all valid modality scenarios, performance is evaluated using Intersection over Union (IoU) and $F_1$ score. Since there is only one selected class in each binary segmentation task, we report these metrics with respect to the corresponding target class.

\subsection{Experimental results and discussion}

\begin{figure*}[t]
\centering
\includegraphics[width=\textwidth]{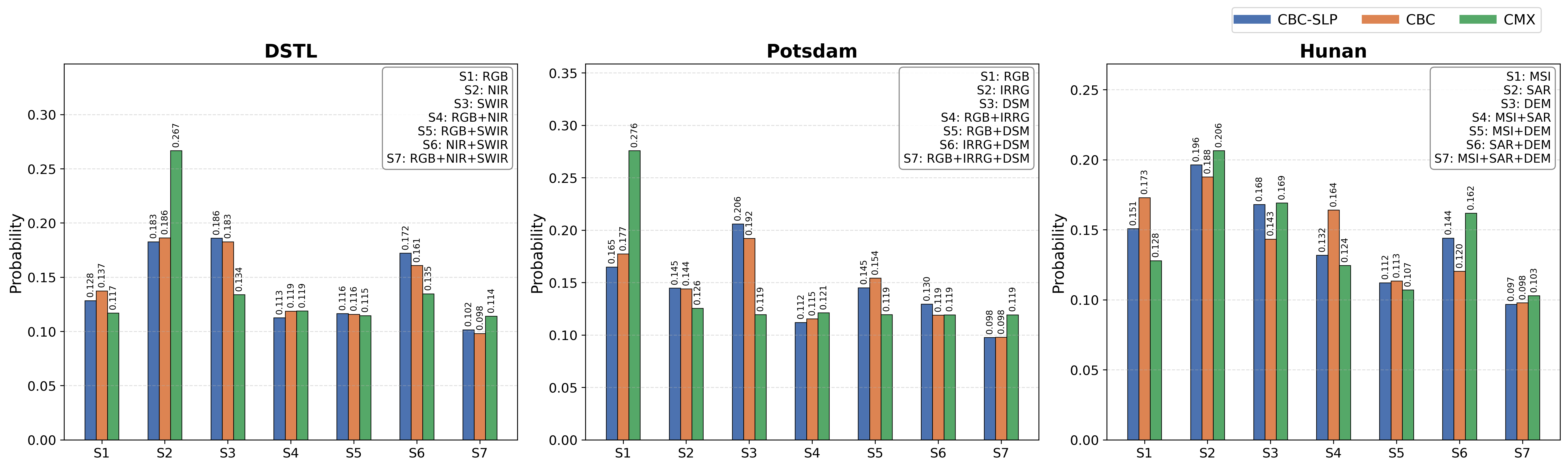}
\caption{Scenario probability distributions obtained for the DSTL, Potsdam, and Hunan image sets.}
\label{fig:probabilities}
\end{figure*}

Fig.~\ref{fig:probabilities} shows the scenario probability distributions learned for DSTL, Potsdam, and Hunan image sets. Different non-uniform distributions are observed across image sets, which demonstrates that modality availability scenarios do not disturb the pretrained latent space to the same extent.

\begin{table*}[t]
\centering
\caption{DSTL test set results for CBC-SLP model under different fine-tuning strategies}
\label{tab:dstl_crop_finetune}
\setlength{\tabcolsep}{3.6pt}
\renewcommand{\arraystretch}{1.15}
\resizebox{\textwidth}{!}{%
\begin{tabular}{ccc cc cc cc cc}
\toprule
\multicolumn{3}{c}{\textbf{Modalities}} &
\multicolumn{2}{c}{\textbf{CBC-SLP (baseline)}} &
\multicolumn{2}{c}{\textbf{CBC-SLP}} &
\multicolumn{2}{c}{\textbf{CBC-SLP}} &
\multicolumn{2}{c}{\textbf{CBC-SLP}} \\
\cmidrule(lr){1-3}\cmidrule(lr){4-5}\cmidrule(lr){6-7}\cmidrule(lr){8-9}\cmidrule(lr){10-11}
\textbf{RGB} & \textbf{NIR} & \textbf{SWIR} &
\multicolumn{2}{c}{\makecell{\textbf{No}\\\textbf{fine-tuning}}} &
\multicolumn{2}{c}{\makecell{\textbf{Fine-tune with random}\\\textbf{modality dropout}}} &
\multicolumn{2}{c}{\makecell{\textbf{Fine-tune with LoRA-based}\\\textbf{adaptation}}} &
\multicolumn{2}{c}{\makecell{\textbf{Fine-tune with}\\\textbf{our strategy}}} \\
\cmidrule(lr){4-5}\cmidrule(lr){6-7}\cmidrule(lr){8-9}\cmidrule(lr){10-11}
 &  &  &
\textbf{IoU} & \textbf{F$_1$} &
\textbf{IoU} & \textbf{F$_1$} &
\textbf{IoU} & \textbf{F$_1$} &
\textbf{IoU} & \textbf{F$_1$} \\
\midrule
\cmark & \cmark & \cmark &
0.903 $\pm$ 0.255 & 0.916 $\pm$ 0.242 &
0.904 $\pm$ 0.253 & 0.917 $\pm$ 0.240 &
0.903 $\pm$ 0.255 & 0.916 $\pm$ 0.242 &
\textbf{0.907 $\pm$ 0.248} & \textbf{0.921 $\pm$ 0.234} \\

\xmark & \cmark & \cmark &
0.901 $\pm$ 0.256 & 0.915 $\pm$ 0.241 &
0.903 $\pm$ 0.254 & 0.917 $\pm$ 0.241 &
0.901 $\pm$ 0.257 & 0.915 $\pm$ 0.242 &
\textbf{0.904 $\pm$ 0.250} & \textbf{0.919 $\pm$ 0.236} \\

\cmark & \xmark & \cmark &
0.893 $\pm$ 0.269 & 0.907 $\pm$ 0.255 &
0.890 $\pm$ 0.267 & 0.906 $\pm$ 0.250 &
0.893 $\pm$ 0.268 & 0.908 $\pm$ 0.255 &
\textbf{0.894 $\pm$ 0.264} & \textbf{0.910 $\pm$ 0.247} \\

\cmark & \cmark & \xmark &
0.902 $\pm$ 0.256 & 0.916 $\pm$ 0.241 &
0.903 $\pm$ 0.254 & 0.917 $\pm$ 0.240 &
0.902 $\pm$ 0.256 & 0.916 $\pm$ 0.241 &
\textbf{0.905 $\pm$ 0.250} & \textbf{0.919 $\pm$ 0.234} \\

\cmark & \xmark & \xmark &
0.876 $\pm$ 0.300 & 0.887 $\pm$ 0.293 &
0.874 $\pm$ 0.299 & 0.886 $\pm$ 0.289 &
0.876 $\pm$ 0.300 & 0.887 $\pm$ 0.293 &
\textbf{0.877 $\pm$ 0.297} & \textbf{0.888 $\pm$ 0.289} \\

\xmark & \cmark & \xmark &
0.896 $\pm$ 0.265 & 0.910 $\pm$ 0.251 &
0.899 $\pm$ 0.264 & 0.912 $\pm$ 0.252 &
0.896 $\pm$ 0.265 & 0.910 $\pm$ 0.251 &
\textbf{0.900 $\pm$ 0.259} & \textbf{0.914 $\pm$ 0.245} \\

\xmark & \xmark & \cmark &
0.865 $\pm$ 0.306 & \textbf{0.879 $\pm$ 0.294} &
0.863 $\pm$ 0.314 & 0.875 $\pm$ 0.304 &
0.865 $\pm$ 0.306 & 0.879 $\pm$ 0.295 &
\textbf{0.865 $\pm$ 0.306} & 0.877 $\pm$ 0.302 \\
\bottomrule
\end{tabular}
}

\vspace{1mm}
\footnotesize
\emph{Note:} The best scores under each modality availability combination are marked as bold.
\xmark\ and \cmark\ are used to represent missing and available modalities in the DSTL image set, respectively. Crop class is used.
\end{table*}

\begin{table*}[h]
\centering
\caption{Potsdam test set results for CBC-SLP model under different fine-tuning strategies}
\label{tab:potsdam_tree_finetune}
\setlength{\tabcolsep}{3.6pt}
\renewcommand{\arraystretch}{1.15}
\resizebox{\textwidth}{!}{%
\begin{tabular}{ccc cc cc cc cc}
\toprule
\multicolumn{3}{c}{\textbf{Modalities}} &
\multicolumn{2}{c}{\textbf{CBC-SLP (baseline)}} &
\multicolumn{2}{c}{\textbf{CBC-SLP}} &
\multicolumn{2}{c}{\textbf{CBC-SLP}} &
\multicolumn{2}{c}{\textbf{CBC-SLP}} \\
\cmidrule(lr){1-3}\cmidrule(lr){4-5}\cmidrule(lr){6-7}\cmidrule(lr){8-9}\cmidrule(lr){10-11}
\textbf{RGB} & \textbf{IRRG} & \textbf{DSM} &
\multicolumn{2}{c}{\makecell{\textbf{No}\\\textbf{fine-tuning}}} &
\multicolumn{2}{c}{\makecell{\textbf{Fine-tune with random}\\\textbf{modality dropout}}} &
\multicolumn{2}{c}{\makecell{\textbf{Fine-tune with LoRA-based}\\\textbf{adaptation}}} &
\multicolumn{2}{c}{\makecell{\textbf{Fine-tune with}\\\textbf{our strategy}}} \\
\cmidrule(lr){4-5}\cmidrule(lr){6-7}\cmidrule(lr){8-9}\cmidrule(lr){10-11}
 &  &  &
\textbf{IoU} & \textbf{F$_1$} &
\textbf{IoU} & \textbf{F$_1$} &
\textbf{IoU} & \textbf{F$_1$} &
\textbf{IoU} & \textbf{F$_1$} \\
\midrule
\cmark & \cmark & \cmark &
0.766 $\pm$ 0.325 & 0.811 $\pm$ 0.312 &
0.774 $\pm$ 0.320 & 0.818 $\pm$ 0.307 &
0.768 $\pm$ 0.324 & 0.812 $\pm$ 0.310 &
\textbf{0.776 $\pm$ 0.318} & \textbf{0.819 $\pm$ 0.305} \\

\xmark & \cmark & \cmark &
0.758 $\pm$ 0.331 & 0.803 $\pm$ 0.317 &
0.767 $\pm$ 0.324 & 0.812 $\pm$ 0.311 &
0.760 $\pm$ 0.329 & 0.805 $\pm$ 0.315 &
\textbf{0.768 $\pm$ 0.323} & \textbf{0.813 $\pm$ 0.309} \\

\cmark & \xmark & \cmark &
0.711 $\pm$ 0.361 & 0.758 $\pm$ 0.349 &
0.724 $\pm$ 0.354 & 0.770 $\pm$ 0.342 &
0.714 $\pm$ 0.359 & 0.761 $\pm$ 0.347 &
\textbf{0.730 $\pm$ 0.351} & \textbf{0.777 $\pm$ 0.339} \\

\cmark & \cmark & \xmark &
0.761 $\pm$ 0.329 & 0.806 $\pm$ 0.316 &
0.769 $\pm$ 0.325 & 0.813 $\pm$ 0.312 &
0.761 $\pm$ 0.329 & 0.806 $\pm$ 0.316 &
\textbf{0.771 $\pm$ 0.322} & \textbf{0.815 $\pm$ 0.309} \\

\cmark & \xmark & \xmark &
0.691 $\pm$ 0.376 & 0.737 $\pm$ 0.366 &
0.704 $\pm$ 0.369 & 0.749 $\pm$ 0.360 &
0.691 $\pm$ 0.377 & 0.736 $\pm$ 0.367 &
\textbf{0.713 $\pm$ 0.363} & \textbf{0.758 $\pm$ 0.352} \\

\xmark & \cmark & \xmark &
0.750 $\pm$ 0.337 & 0.795 $\pm$ 0.324 &
0.760 $\pm$ 0.330 & 0.805 $\pm$ 0.317 &
0.749 $\pm$ 0.337 & 0.795 $\pm$ 0.324 &
\textbf{0.761 $\pm$ 0.329} & \textbf{0.806 $\pm$ 0.316} \\

\xmark & \xmark & \cmark &
0.576 $\pm$ 0.432 & 0.614 $\pm$ 0.427 &
0.583 $\pm$ 0.431 & 0.622 $\pm$ 0.427 &
\textbf{0.583 $\pm$ 0.429} & \textbf{0.622 $\pm$ 0.425} &
0.579 $\pm$ 0.434 & 0.616 $\pm$ 0.429 \\
\bottomrule
\end{tabular}
}

\vspace{1mm}
\footnotesize
\emph{Note:} The best scores under each modality availability combination are marked as bold.
\xmark\ and \cmark\ are used to represent missing and available modalities in the Potsdam image set, respectively. Tree class is used.

\end{table*}

\begin{table*}[h]
\centering
\caption{Hunan test set results for CBC-SLP model under different fine-tuning strategies}
\label{tab:hunan_forest_finetune}
\setlength{\tabcolsep}{3.6pt}
\renewcommand{\arraystretch}{1.15}
\resizebox{\textwidth}{!}{%
\begin{tabular}{ccc cc cc cc cc}
\toprule
\multicolumn{3}{c}{\textbf{Modalities}} &
\multicolumn{2}{c}{\textbf{CBC-SLP (baseline)}} &
\multicolumn{2}{c}{\textbf{CBC-SLP}} &
\multicolumn{2}{c}{\textbf{CBC-SLP}} &
\multicolumn{2}{c}{\textbf{CBC-SLP}} \\
\cmidrule(lr){1-3}\cmidrule(lr){4-5}\cmidrule(lr){6-7}\cmidrule(lr){8-9}\cmidrule(lr){10-11}
\textbf{SAR} & \textbf{MSI} & \textbf{DEM} &
\multicolumn{2}{c}{\makecell{\textbf{No}\\\textbf{fine-tuning}}} &
\multicolumn{2}{c}{\makecell{\textbf{Fine-tune with random}\\\textbf{modality dropout}}} &
\multicolumn{2}{c}{\makecell{\textbf{Fine-tune with LoRA-based}\\\textbf{adaptation}}} &
\multicolumn{2}{c}{\makecell{\textbf{Fine-tune with}\\\textbf{our strategy}}} \\
\cmidrule(lr){4-5}\cmidrule(lr){6-7}\cmidrule(lr){8-9}\cmidrule(lr){10-11}
 &  &  &
\textbf{IoU} & \textbf{F$_1$} &
\textbf{IoU} & \textbf{F$_1$} &
\textbf{IoU} & \textbf{F$_1$} &
\textbf{IoU} & \textbf{F$_1$} \\
\midrule
\cmark & \cmark & \cmark &
0.659 $\pm$ 0.299 & 0.743 $\pm$ 0.287 &
0.670 $\pm$ 0.290 & 0.755 $\pm$ 0.275 &
0.659 $\pm$ 0.299 & 0.744 $\pm$ 0.286 &
\textbf{0.672 $\pm$ 0.289} & \textbf{0.757 $\pm$ 0.269} \\

\xmark & \cmark & \cmark &
0.640 $\pm$ 0.304 & 0.727 $\pm$ 0.295 &
0.642 $\pm$ 0.305 & 0.728 $\pm$ 0.297 &
0.640 $\pm$ 0.304 & 0.727 $\pm$ 0.295 &
\textbf{0.645 $\pm$ 0.302} & \textbf{0.731 $\pm$ 0.290} \\

\cmark & \xmark & \cmark &
0.582 $\pm$ 0.323 & 0.670 $\pm$ 0.326 &
0.591 $\pm$ 0.318 & 0.681 $\pm$ 0.318 &
0.583 $\pm$ 0.323 & 0.671 $\pm$ 0.326 &
\textbf{0.595 $\pm$ 0.314} & \textbf{0.686 $\pm$ 0.313} \\

\cmark & \cmark & \xmark &
0.652 $\pm$ 0.300 & 0.737 $\pm$ 0.291 &
0.659 $\pm$ 0.298 & 0.743 $\pm$ 0.290 &
0.653 $\pm$ 0.300 & 0.738 $\pm$ 0.290 &
\textbf{0.663 $\pm$ 0.297} & \textbf{0.747 $\pm$ 0.286} \\

\cmark & \xmark & \xmark &
0.569 $\pm$ 0.298 & \textbf{0.668 $\pm$ 0.302} &
0.561 $\pm$ 0.313 & 0.656 $\pm$ 0.318 &
0.558 $\pm$ 0.316 & 0.652 $\pm$ 0.323 &
\textbf{0.569 $\pm$ 0.297} & 0.663 $\pm$ 0.312 \\

\xmark & \cmark & \xmark &
0.616 $\pm$ 0.313 & 0.704 $\pm$ 0.306 &
0.620 $\pm$ 0.313 & 0.707 $\pm$ 0.305 &
0.617 $\pm$ 0.313 & 0.705 $\pm$ 0.305 &
\textbf{0.626 $\pm$ 0.310} & \textbf{0.713 $\pm$ 0.302} \\

\xmark & \xmark & \cmark &
\textbf{0.518 $\pm$ 0.322} & \textbf{0.616 $\pm$ 0.321} &
0.515 $\pm$ 0.345 & 0.601 $\pm$ 0.353 &
0.504 $\pm$ 0.344 & 0.592 $\pm$ 0.351 &
0.509 $\pm$ 0.339 & 0.598 $\pm$ 0.345 \\
\bottomrule
\end{tabular}
}
\vspace{1mm}
\footnotesize
\emph{Note:} The best scores under each modality availability combination are marked as bold.
\xmark\ and \cmark\ are used to represent missing and available modalities in the Hunan image set, respectively. Forest class is used.
\end{table*}

\begin{table*}[h]
\centering
\caption{DSTL test set results for CBC model under different fine-tuning strategies}
\label{tab:dstl_crop_finetune_cbc}
\setlength{\tabcolsep}{3.6pt}
\renewcommand{\arraystretch}{1.15}
\resizebox{\textwidth}{!}{%
\begin{tabular}{ccc cc cc cc cc}
\toprule
\multicolumn{3}{c}{\textbf{Modalities}} &
\multicolumn{2}{c}{\textbf{CBC (baseline)}} &
\multicolumn{2}{c}{\textbf{CBC}} &
\multicolumn{2}{c}{\textbf{CBC}} &
\multicolumn{2}{c}{\textbf{CBC}} \\
\cmidrule(lr){1-3}\cmidrule(lr){4-5}\cmidrule(lr){6-7}\cmidrule(lr){8-9}\cmidrule(lr){10-11}
\textbf{RGB} & \textbf{NIR} & \textbf{SWIR} &
\multicolumn{2}{c}{\makecell{\textbf{No}\\\textbf{fine-tuning}}} &
\multicolumn{2}{c}{\makecell{\textbf{Fine-tune with random}\\\textbf{modality dropout}}} &
\multicolumn{2}{c}{\makecell{\textbf{Fine-tune with LoRA-based}\\\textbf{adaptation}}} &
\multicolumn{2}{c}{\makecell{\textbf{Fine-tune with}\\\textbf{our strategy}}} \\
\cmidrule(lr){4-5}\cmidrule(lr){6-7}\cmidrule(lr){8-9}\cmidrule(lr){10-11}
 &  &  &
\textbf{IoU} & \textbf{F$_1$} &
\textbf{IoU} & \textbf{F$_1$} &
\textbf{IoU} & \textbf{F$_1$} &
\textbf{IoU} & \textbf{F$_1$} \\
\midrule
\cmark & \cmark & \cmark &
0.881 $\pm$ 0.289 & 0.894 $\pm$ 0.279 &
0.885 $\pm$ 0.282 & 0.899 $\pm$ 0.270 &
0.881 $\pm$ 0.289 & 0.894 $\pm$ 0.279 &
\textbf{0.885 $\pm$ 0.281} & 0.899 $\pm$ 0.269 \\

\xmark & \cmark & \cmark &
0.880 $\pm$ 0.292 & 0.892 $\pm$ 0.284 &
0.885 $\pm$ 0.281 & 0.899 $\pm$ 0.271 &
0.880 $\pm$ 0.292 & 0.892 $\pm$ 0.284 &
\textbf{0.886 $\pm$ 0.281} & 0.899 $\pm$ 0.286 \\

\cmark & \xmark & \cmark &
0.871 $\pm$ 0.299 & 0.885 $\pm$ 0.289 &
0.869 $\pm$ 0.303 & 0.882 $\pm$ 0.293 &
\textbf{0.871 $\pm$ 0.299} & \textbf{0.885 $\pm$ 0.289} &
0.869 $\pm$ 0.302 & 0.882 $\pm$ 0.293 \\

\cmark & \cmark & \xmark &
0.881 $\pm$ 0.291 & 0.893 $\pm$ 0.282 &
0.884 $\pm$ 0.285 & 0.897 $\pm$ 0.275 &
0.881 $\pm$ 0.291 & 0.893 $\pm$ 0.282 &
\textbf{0.884 $\pm$ 0.284} & 0.897 $\pm$ 0.272 \\

\cmark & \xmark & \xmark &
0.872 $\pm$ 0.298 & 0.886 $\pm$ 0.287 &
0.868 $\pm$ 0.301 & 0.883 $\pm$ 0.291 &
0.872 $\pm$ 0.298 & 0.886 $\pm$ 0.287 &
\textbf{0.876 $\pm$ 0.296} & \textbf{0.889 $\pm$ 0.286} \\

\xmark & \cmark & \xmark &
0.879 $\pm$ 0.294 & 0.891 $\pm$ 0.286 &
0.883 $\pm$ 0.290 & 0.894 $\pm$ 0.282 &
0.879 $\pm$ 0.294 & 0.891 $\pm$ 0.286 &
\textbf{0.885 $\pm$ 0.287} & 0.897 $\pm$ 0.277 \\

\xmark & \xmark & \cmark &
0.856 $\pm$ 0.323 & 0.868 $\pm$ 0.313 &
0.860 $\pm$ 0.314 & 0.874 $\pm$ 0.303 &
0.856 $\pm$ 0.323 & 0.868 $\pm$ 0.313 &
\textbf{0.862 $\pm$ 0.313} & \textbf{0.875 $\pm$ 0.302} \\
\bottomrule
\end{tabular}
}
\vspace{1mm}
\footnotesize
\emph{Note:} The best scores under each modality availability combination are marked as bold.
\xmark\ and \cmark\ are used to represent missing and available modalities in the DSTL image set, respectively. Crop class is used.
\end{table*}

\begin{table*}[h]
\centering
\caption{Potsdam test set results for CBC model under different fine-tuning strategies}
\label{tab:potsdam_tree_finetune_cbc}
\setlength{\tabcolsep}{3.6pt}
\renewcommand{\arraystretch}{1.15}
\resizebox{\textwidth}{!}{%
\begin{tabular}{ccc cc cc cc cc}
\toprule
\multicolumn{3}{c}{\textbf{Modalities}} &
\multicolumn{2}{c}{\textbf{CBC (baseline)}} &
\multicolumn{2}{c}{\textbf{CBC}} &
\multicolumn{2}{c}{\textbf{CBC}} &
\multicolumn{2}{c}{\textbf{CBC}} \\
\cmidrule(lr){1-3}\cmidrule(lr){4-5}\cmidrule(lr){6-7}\cmidrule(lr){8-9}\cmidrule(lr){10-11}
\textbf{RGB} & \textbf{IRRG} & \textbf{DSM} &
\multicolumn{2}{c}{\makecell{\textbf{No}\\\textbf{fine-tuning}}} &
\multicolumn{2}{c}{\makecell{\textbf{Fine-tune with random}\\\textbf{modality dropout}}} &
\multicolumn{2}{c}{\makecell{\textbf{Fine-tune with LoRA-based}\\\textbf{adaptation}}} &
\multicolumn{2}{c}{\makecell{\textbf{Fine-tune with}\\\textbf{our strategy}}} \\
\cmidrule(lr){4-5}\cmidrule(lr){6-7}\cmidrule(lr){8-9}\cmidrule(lr){10-11}
 &  &  &
\textbf{IoU} & \textbf{F$_1$} &
\textbf{IoU} & \textbf{F$_1$} &
\textbf{IoU} & \textbf{F$_1$} &
\textbf{IoU} & \textbf{F$_1$} \\
\midrule
\cmark & \cmark & \cmark &
0.762 $\pm$ 0.331 & 0.805 $\pm$ 0.320 &
0.772 $\pm$ 0.323 & 0.812 $\pm$ 0.311 &
0.762 $\pm$ 0.332 & 0.805 $\pm$ 0.321 &
\textbf{0.776 $\pm$ 0.320} & \textbf{0.815 $\pm$ 0.320} \\

\xmark & \cmark & \cmark &
0.758 $\pm$ 0.332 & 0.802 $\pm$ 0.321 &
0.765 $\pm$ 0.325 & 0.810 $\pm$ 0.312 &
0.757 $\pm$ 0.333 & 0.802 $\pm$ 0.322 &
\textbf{0.767 $\pm$ 0.326} & \textbf{0.811 $\pm$ 0.313} \\

\cmark & \xmark & \cmark &
0.713 $\pm$ 0.362 & 0.759 $\pm$ 0.351 &
0.718 $\pm$ 0.354 & 0.766 $\pm$ 0.341 &
0.712 $\pm$ 0.363 & 0.757 $\pm$ 0.353 &
\textbf{0.720 $\pm$ 0.357} & \textbf{0.766 $\pm$ 0.346} \\

\cmark & \cmark & \xmark &
0.753 $\pm$ 0.339 & 0.796 $\pm$ 0.329 &
\textbf{0.766 $\pm$ 0.328} & \textbf{0.809 $\pm$ 0.316} &
0.752 $\pm$ 0.339 & 0.795 $\pm$ 0.329 &
0.762 $\pm$ 0.330 & 0.806 $\pm$ 0.318 \\

\cmark & \xmark & \xmark &
0.682 $\pm$ 0.386 & 0.725 $\pm$ 0.379 &
\textbf{0.698 $\pm$ 0.375} & \textbf{0.741 $\pm$ 0.367} &
0.682 $\pm$ 0.386 & 0.724 $\pm$ 0.379 &
0.680 $\pm$ 0.383 & 0.725 $\pm$ 0.375 \\

\xmark & \cmark & \xmark &
0.745 $\pm$ 0.343 & 0.789 $\pm$ 0.333 &
\textbf{0.759 $\pm$ 0.332} & \textbf{0.803 $\pm$ 0.320} &
0.745 $\pm$ 0.344 & 0.789 $\pm$ 0.333 &
0.748 $\pm$ 0.338 & 0.793 $\pm$ 0.327 \\

\xmark & \xmark & \cmark &
0.584 $\pm$ 0.420 & 0.628 $\pm$ 0.415 &
0.589 $\pm$ 0.424 & 0.630 $\pm$ 0.419 &
0.583 $\pm$ 0.423 & 0.625 $\pm$ 0.418 &
\textbf{0.591 $\pm$ 0.422} & \textbf{0.634 $\pm$ 0.416} \\
\bottomrule
\end{tabular}
}
\vspace{1mm}
\footnotesize
\emph{Note:} The best scores under each modality availability combination are marked as bold.
\xmark\ and \cmark\ are used to represent missing and available modalities in the Potsdam image set, respectively. Tree class is used.
\end{table*}

\begin{table*}[h]
\centering
\caption{Hunan test set results for CBC model under different fine-tuning strategies}
\label{tab:hunan_forest_finetune_cbc}
\setlength{\tabcolsep}{3.6pt}
\renewcommand{\arraystretch}{1.15}
\resizebox{\textwidth}{!}{%
\begin{tabular}{ccc cc cc cc cc}
\toprule
\multicolumn{3}{c}{\textbf{Modalities}} &
\multicolumn{2}{c}{\textbf{CBC (baseline)}} &
\multicolumn{2}{c}{\textbf{CBC}} &
\multicolumn{2}{c}{\textbf{CBC}} &
\multicolumn{2}{c}{\textbf{CBC}} \\
\cmidrule(lr){1-3}\cmidrule(lr){4-5}\cmidrule(lr){6-7}\cmidrule(lr){8-9}\cmidrule(lr){10-11}
\textbf{SAR} & \textbf{MSI} & \textbf{DEM} &
\multicolumn{2}{c}{\makecell{\textbf{No}\\\textbf{fine-tuning}}} &
\multicolumn{2}{c}{\makecell{\textbf{Fine-tune with random}\\\textbf{modality dropout}}} &
\multicolumn{2}{c}{\makecell{\textbf{Fine-tune with LoRA-based}\\\textbf{adaptation}}} &
\multicolumn{2}{c}{\makecell{\textbf{Fine-tune with}\\\textbf{our strategy}}} \\
\cmidrule(lr){4-5}\cmidrule(lr){6-7}\cmidrule(lr){8-9}\cmidrule(lr){10-11}
 &  &  &
\textbf{IoU} & \textbf{F$_1$} &
\textbf{IoU} & \textbf{F$_1$} &
\textbf{IoU} & \textbf{F$_1$} &
\textbf{IoU} & \textbf{F$_1$} \\
\midrule
\cmark & \cmark & \cmark &
0.621 $\pm$ 0.309 & 0.709 $\pm$ 0.300 &
0.654 $\pm$ 0.299 & 0.739 $\pm$ 0.288 &
0.620 $\pm$ 0.309 & 0.708 $\pm$ 0.301 &
\textbf{0.658 $\pm$ 0.299} & \textbf{0.742 $\pm$ 0.289} \\

\xmark & \cmark & \cmark &
0.606 $\pm$ 0.314 & 0.696 $\pm$ 0.304 &
0.633 $\pm$ 0.306 & 0.721 $\pm$ 0.295 &
0.606 $\pm$ 0.314 & 0.696 $\pm$ 0.304 &
\textbf{0.640 $\pm$ 0.302} & \textbf{0.728 $\pm$ 0.291} \\

\cmark & \xmark & \cmark &
0.526 $\pm$ 0.319 & 0.623 $\pm$ 0.323 &
0.558 $\pm$ 0.307 & 0.656 $\pm$ 0.309 &
0.526 $\pm$ 0.319 & 0.623 $\pm$ 0.322 &
\textbf{0.564 $\pm$ 0.304} & \textbf{0.663 $\pm$ 0.306} \\

\cmark & \cmark & \xmark &
0.604 $\pm$ 0.321 & 0.690 $\pm$ 0.317 &
0.636 $\pm$ 0.313 & 0.720 $\pm$ 0.305 &
0.603 $\pm$ 0.321 & 0.690 $\pm$ 0.316 &
\textbf{0.641 $\pm$ 0.313} & \textbf{0.724 $\pm$ 0.306} \\

\cmark & \xmark & \xmark &
0.497 $\pm$ 0.315 & 0.599 $\pm$ 0.318 &
0.526 $\pm$ 0.312 & 0.625 $\pm$ 0.320 &
0.498 $\pm$ 0.315 & 0.599 $\pm$ 0.318 &
\textbf{0.532 $\pm$ 0.310} & \textbf{0.631 $\pm$ 0.318} \\

\xmark & \cmark & \xmark &
0.590 $\pm$ 0.324 & 0.679 $\pm$ 0.316 &
0.620 $\pm$ 0.312 & 0.708 $\pm$ 0.301 &
0.591 $\pm$ 0.324 & 0.679 $\pm$ 0.316 &
\textbf{0.626 $\pm$ 0.309} & \textbf{0.714 $\pm$ 0.299} \\

\xmark & \xmark & \cmark &
0.517 $\pm$ 0.345 & 0.603 $\pm$ 0.350 &
0.537 $\pm$ 0.342 & 0.622 $\pm$ 0.349 &
0.517 $\pm$ 0.345 & 0.603 $\pm$ 0.350 &
\textbf{0.545 $\pm$ 0.343} & \textbf{0.630 $\pm$ 0.348} \\
\bottomrule
\end{tabular}
}
\vspace{1mm}
\footnotesize
\emph{Note:} The best scores under each modality availability combination are marked as bold.
\xmark\ and \cmark\ are used to represent missing and available modalities in the Hunan image set, respectively. Forest class is used.
\end{table*}

\begin{table*}[h]
\centering
\caption{DSTL test set results for CMX model under different fine-tuning strategies}
\label{tab:dstl_crop_finetune_cmx}
\setlength{\tabcolsep}{3.6pt}
\renewcommand{\arraystretch}{1.15}
\resizebox{\textwidth}{!}{%
\begin{tabular}{ccc cc cc cc cc}
\toprule
\multicolumn{3}{c}{\textbf{Modalities}} &
\multicolumn{2}{c}{\textbf{CMX (baseline)}} &
\multicolumn{2}{c}{\textbf{CMX}} &
\multicolumn{2}{c}{\textbf{CMX}} &
\multicolumn{2}{c}{\textbf{CMX}} \\
\cmidrule(lr){1-3}\cmidrule(lr){4-5}\cmidrule(lr){6-7}\cmidrule(lr){8-9}\cmidrule(lr){10-11}
\textbf{RGB} & \textbf{NIR} & \textbf{SWIR} &
\multicolumn{2}{c}{\makecell{\textbf{No}\\\textbf{fine-tuning}}} &
\multicolumn{2}{c}{\makecell{\textbf{Fine-tune with random}\\\textbf{modality dropout}}} &
\multicolumn{2}{c}{\makecell{\textbf{Fine-tune with LoRA-based}\\\textbf{adaptation}}} &
\multicolumn{2}{c}{\makecell{\textbf{Fine-tune with}\\\textbf{our strategy}}} \\
\cmidrule(lr){4-5}\cmidrule(lr){6-7}\cmidrule(lr){8-9}\cmidrule(lr){10-11}
 &  &  &
\textbf{IoU} & \textbf{F$_1$} &
\textbf{IoU} & \textbf{F$_1$} &
\textbf{IoU} & \textbf{F$_1$} &
\textbf{IoU} & \textbf{F$_1$} \\
\midrule
\cmark & \cmark & \cmark &
0.860 $\pm$ 0.324 & 0.869 $\pm$ 0.319 &
0.863 $\pm$ 0.321 & 0.872 $\pm$ 0.316 &
0.860 $\pm$ 0.324 & 0.869 $\pm$ 0.319 &
\textbf{0.864 $\pm$ 0.320} & \textbf{0.872 $\pm$ 0.315} \\

\xmark & \cmark & \cmark &
0.853 $\pm$ 0.332 & 0.862 $\pm$ 0.326 &
0.853 $\pm$ 0.330 & 0.863 $\pm$ 0.324 &
0.852 $\pm$ 0.335 & 0.860 $\pm$ 0.329 &
\textbf{0.856 $\pm$ 0.325} & \textbf{0.867 $\pm$ 0.318} \\

\cmark & \xmark & \cmark &
0.859 $\pm$ 0.325 & 0.868 $\pm$ 0.320 &
0.862 $\pm$ 0.322 & 0.871 $\pm$ 0.317 &
0.859 $\pm$ 0.325 & 0.868 $\pm$ 0.320 &
\textbf{0.863 $\pm$ 0.321} & \textbf{0.872 $\pm$ 0.316} \\

\cmark & \cmark & \xmark &
0.853 $\pm$ 0.333 & 0.862 $\pm$ 0.328 &
0.858 $\pm$ 0.327 & 0.867 $\pm$ 0.322 &
0.854 $\pm$ 0.332 & 0.862 $\pm$ 0.328 &
\textbf{0.859 $\pm$ 0.324} & \textbf{0.868 $\pm$ 0.319} \\

\cmark & \xmark & \xmark &
0.854 $\pm$ 0.331 & 0.863 $\pm$ 0.325 &
0.858 $\pm$ 0.326 & 0.867 $\pm$ 0.321 &
0.855 $\pm$ 0.330 & 0.864 $\pm$ 0.325 &
\textbf{0.860 $\pm$ 0.324} & \textbf{0.869 $\pm$ 0.320} \\

\xmark & \cmark & \xmark &
0.806 $\pm$ 0.377 & 0.815 $\pm$ 0.371 &
0.827 $\pm$ 0.348 & 0.841 $\pm$ 0.339 &
0.805 $\pm$ 0.378 & 0.814 $\pm$ 0.373 &
\textbf{0.830 $\pm$ 0.348} & \textbf{0.842 $\pm$ 0.341} \\

\xmark & \xmark & \cmark &
0.849 $\pm$ 0.335 & 0.859 $\pm$ 0.328 &
0.850 $\pm$ 0.333 & 0.860 $\pm$ 0.326 &
0.847 $\pm$ 0.338 & 0.856 $\pm$ 0.332 &
\textbf{0.853 $\pm$ 0.327} & \textbf{0.864 $\pm$ 0.319} \\
\bottomrule
\end{tabular}
}
\vspace{1mm}
\footnotesize
\emph{Note:} The best scores under each modality availability combination are marked as bold.
\xmark\ and \cmark\ are used to represent missing and available modalities in the DSTL image set, respectively. Crop class is used.
\end{table*}

\begin{table*}[h]
\centering
\caption{Potsdam test set results for CMX model under different fine-tuning strategies}
\label{tab:potsdam_tree_finetune_cmx}
\setlength{\tabcolsep}{3.6pt}
\renewcommand{\arraystretch}{1.15}
\resizebox{\textwidth}{!}{%
\begin{tabular}{ccc cc cc cc cc}
\toprule
\multicolumn{3}{c}{\textbf{Modalities}} &
\multicolumn{2}{c}{\textbf{CMX (baseline)}} &
\multicolumn{2}{c}{\textbf{CMX}} &
\multicolumn{2}{c}{\textbf{CMX}} &
\multicolumn{2}{c}{\textbf{CMX}} \\
\cmidrule(lr){1-3}\cmidrule(lr){4-5}\cmidrule(lr){6-7}\cmidrule(lr){8-9}\cmidrule(lr){10-11}
\textbf{RGB} & \textbf{IRRG} & \textbf{DSM} &
\multicolumn{2}{c}{\makecell{\textbf{No}\\\textbf{fine-tuning}}} &
\multicolumn{2}{c}{\makecell{\textbf{Fine-tune with random}\\\textbf{modality dropout}}} &
\multicolumn{2}{c}{\makecell{\textbf{Fine-tune with LoRA-based}\\\textbf{adaptation}}} &
\multicolumn{2}{c}{\makecell{\textbf{Fine-tune with}\\\textbf{our strategy}}} \\
\cmidrule(lr){4-5}\cmidrule(lr){6-7}\cmidrule(lr){8-9}\cmidrule(lr){10-11}
 &  &  &
\textbf{IoU} & \textbf{F$_1$} &
\textbf{IoU} & \textbf{F$_1$} &
\textbf{IoU} & \textbf{F$_1$} &
\textbf{IoU} & \textbf{F$_1$} \\
\midrule
\cmark & \cmark & \cmark &
0.541 $\pm$ 0.426 & 0.590 $\pm$ 0.408 &
0.561 $\pm$ 0.416 & 0.614 $\pm$ 0.396 &
0.538 $\pm$ 0.431 & 0.585 $\pm$ 0.413 &
\textbf{0.574 $\pm$ 0.406} & \textbf{0.631 $\pm$ 0.385} \\

\xmark & \cmark & \cmark &
0.532 $\pm$ 0.434 & 0.577 $\pm$ 0.418 &
0.535 $\pm$ 0.436 & 0.579 $\pm$ 0.421 &
0.526 $\pm$ 0.440 & 0.568 $\pm$ 0.425 &
\textbf{0.562 $\pm$ 0.391} & \textbf{0.629 $\pm$ 0.368} \\

\cmark & \xmark & \cmark &
0.516 $\pm$ 0.448 & 0.553 $\pm$ 0.436 &
\textbf{0.527 $\pm$ 0.441} & \textbf{0.568 $\pm$ 0.428} &
0.526 $\pm$ 0.441 & 0.567 $\pm$ 0.428 &
0.508 $\pm$ 0.452 & 0.543 $\pm$ 0.440 \\

\cmark & \cmark & \xmark &
0.539 $\pm$ 0.426 & 0.589 $\pm$ 0.408 &
0.562 $\pm$ 0.415 & 0.616 $\pm$ 0.395 &
0.536 $\pm$ 0.431 & 0.583 $\pm$ 0.413 &
\textbf{0.573 $\pm$ 0.407} & \textbf{0.630 $\pm$ 0.385} \\

\cmark & \xmark & \xmark &
0.519 $\pm$ 0.445 & 0.558 $\pm$ 0.430 &
0.517 $\pm$ 0.447 & 0.554 $\pm$ 0.434 &
\textbf{0.525 $\pm$ 0.440} & \textbf{0.567 $\pm$ 0.426} &
0.502 $\pm$ 0.455 & 0.535 $\pm$ 0.443 \\

\xmark & \cmark & \xmark &
0.526 $\pm$ 0.435 & 0.571 $\pm$ 0.421 &
0.537 $\pm$ 0.434 & 0.581 $\pm$ 0.419 &
0.522 $\pm$ 0.442 & 0.562 $\pm$ 0.428 &
\textbf{0.561 $\pm$ 0.394} & \textbf{0.626 $\pm$ 0.371} \\

\xmark & \xmark & \cmark &
0.444 $\pm$ 0.492 & 0.448 $\pm$ 0.493 &
0.444 $\pm$ 0.496 & 0.448 $\pm$ 0.496 &
0.444 $\pm$ 0.496 & 0.448 $\pm$ 0.496 &
\textbf{0.444 $\pm$ 0.491} & \textbf{0.448 $\pm$ 0.491} \\
\bottomrule
\end{tabular}
}
\vspace{1mm}
\footnotesize
\emph{Note:} The best scores under each modality availability combination are marked as bold.
\xmark\ and \cmark\ are used to represent missing and available modalities in the Potsdam image set, respectively. Tree class is used.
\end{table*}

\begin{table*}[h]
\centering
\caption{Hunan test set results for CMX model under different fine-tuning strategies}
\label{tab:hunan_forest_finetune_cmx}
\setlength{\tabcolsep}{3.6pt}
\renewcommand{\arraystretch}{1.15}
\resizebox{\textwidth}{!}{%
\begin{tabular}{ccc cc cc cc cc}
\toprule
\multicolumn{3}{c}{\textbf{Modalities}} &
\multicolumn{2}{c}{\textbf{CMX (baseline)}} &
\multicolumn{2}{c}{\textbf{CMX}} &
\multicolumn{2}{c}{\textbf{CMX}} &
\multicolumn{2}{c}{\textbf{CMX}} \\
\cmidrule(lr){1-3}\cmidrule(lr){4-5}\cmidrule(lr){6-7}\cmidrule(lr){8-9}\cmidrule(lr){10-11}
\textbf{SAR} & \textbf{MSI} & \textbf{DEM} &
\multicolumn{2}{c}{\makecell{\textbf{No}\\\textbf{fine-tuning}}} &
\multicolumn{2}{c}{\makecell{\textbf{Fine-tune with random}\\\textbf{modality dropout}}} &
\multicolumn{2}{c}{\makecell{\textbf{Fine-tune with LoRA-based}\\\textbf{adaptation}}} &
\multicolumn{2}{c}{\makecell{\textbf{Fine-tune with}\\\textbf{our strategy}}} \\
\cmidrule(lr){4-5}\cmidrule(lr){6-7}\cmidrule(lr){8-9}\cmidrule(lr){10-11}
 &  &  &
\textbf{IoU} & \textbf{F$_1$} &
\textbf{IoU} & \textbf{F$_1$} &
\textbf{IoU} & \textbf{F$_1$} &
\textbf{IoU} & \textbf{F$_1$} \\
\midrule
\cmark & \cmark & \cmark &
\textbf{0.651 $\pm$ 0.306} & \textbf{0.733 $\pm$ 0.300} &
0.627 $\pm$ 0.321 & 0.709 $\pm$ 0.316 &
0.618 $\pm$ 0.324 & 0.700 $\pm$ 0.322 &
0.634 $\pm$ 0.333 & 0.710 $\pm$ 0.330 \\

\xmark & \cmark & \cmark &
0.625 $\pm$ 0.325 & 0.705 $\pm$ 0.321 &
0.623 $\pm$ 0.321 & 0.706 $\pm$ 0.321 &
0.617 $\pm$ 0.326 & 0.699 $\pm$ 0.322 &
\textbf{0.634 $\pm$ 0.329} & \textbf{0.711 $\pm$ 0.323} \\

\cmark & \xmark & \cmark &
0.555 $\pm$ 0.328 & 0.644 $\pm$ 0.339 &
0.540 $\pm$ 0.330 & 0.630 $\pm$ 0.339 &
0.523 $\pm$ 0.335 & 0.612 $\pm$ 0.348 &
\textbf{0.562 $\pm$ 0.348} & \textbf{0.646 $\pm$ 0.354} \\

\cmark & \cmark & \xmark &
\textbf{0.639 $\pm$ 0.307} & \textbf{0.723 $\pm$ 0.304} &
0.609 $\pm$ 0.322 & 0.694 $\pm$ 0.319 &
0.596 $\pm$ 0.327 & 0.680 $\pm$ 0.326 &
0.607 $\pm$ 0.322 & 0.691 $\pm$ 0.326 \\

\cmark & \xmark & \xmark &
\textbf{0.476 $\pm$ 0.313} & \textbf{0.580 $\pm$ 0.315} &
0.465 $\pm$ 0.330 & 0.563 $\pm$ 0.327 &
0.399 $\pm$ 0.346 & 0.488 $\pm$ 0.339 &
0.441 $\pm$ 0.286 & 0.554 $\pm$ 0.301 \\

\xmark & \cmark & \xmark &
0.599 $\pm$ 0.326 & 0.683 $\pm$ 0.325 &
0.598 $\pm$ 0.323 & 0.684 $\pm$ 0.321 &
0.582 $\pm$ 0.326 & 0.669 $\pm$ 0.325 &
\textbf{0.602 $\pm$ 0.318} & \textbf{0.688 $\pm$ 0.321} \\

\xmark & \xmark & \cmark &
0.509 $\pm$ 0.340 & 0.598 $\pm$ 0.343 &
0.519 $\pm$ 0.339 & 0.606 $\pm$ 0.352 &
0.506 $\pm$ 0.340 & 0.594 $\pm$ 0.352 &
\textbf{0.543 $\pm$ 0.351} & \textbf{0.623 $\pm$ 0.358} \\
\bottomrule
\end{tabular}
}
\vspace{1mm}
\footnotesize
\emph{Note:} The best scores under each modality availability combination are marked as bold.
\xmark\ and \cmark\ are used to represent missing and available modalities in the Hunan image set, respectively. Forest class is used.
\end{table*}

As shown in the results, the proposed training strategy yields consistent improvements across multiple scenario configurations, datasets, and backbones. According to the CBC-SLP results reported in Tables~\ref{tab:dstl_crop_finetune}, \ref{tab:potsdam_tree_finetune}, and \ref{tab:hunan_forest_finetune}, the proposed strategy consistently improves test performance compared to the other training settings. Remarkably, a consistent improvement across all three image sets is observed when all modalities are available, demonstrating that our training strategy does not compromise performance in the complete modality scenario while improving robustness to missing modalities. The largest performance gains in CBC-SLP are observed on image sets with heterogeneous modalities. In Table~\ref{tab:potsdam_tree_finetune}, when IRRG is missing, our latent-space-guided scenario sampling strategy improves IoU by 0.6\% and $F_1$ by 0.7\% compared to fine-tuning with random modality dropout. Specifically, when only RGB is available, the proposed strategy achieves an IoU of 0.713, surpassing the random modality dropout fine-tuning by 0.9\%. In Table~\ref{tab:hunan_forest_finetune}, the proposed strategy yields an IoU gain of 0.8\% over a random modality dropout training policy when only the SAR modality is available. In addition, relative to LoRA-based adaptation, the proposed strategy yields an IoU gain of 1.3\% when all modalities are available and 1.2\% when MSI is missing. These results demonstrate that the proposed fine-tuning policy is particularly effective when the pretrained model already maintains a meaningful structured latent representation, as in the CBC-SLP model.

As shown in Tables~\ref{tab:dstl_crop_finetune_cbc}, \ref{tab:potsdam_tree_finetune_cbc}, \ref{tab:hunan_forest_finetune_cbc}, the proposed strategy consistently benefits the CBC model across the three image sets, with the clearest gains observed on Hunan and more scenario-dependent behavior on Potsdam. Table~\ref{tab:hunan_forest_finetune_cbc} shows that the proposed strategy achieves the best performance on all scenarios, with notable gains for several difficult cases. Specifically, it achieves an IoU of 0.545 when only DEM is available, surpassing the random modality dropout result by 0.8\%. When MSI or DEM is missing, we observe a largest IoU gain of 3.8\% relative to LoRA-based adaptation, demonstrating that latent-space-guided fine-tuning provides more effective adaptation than generic fine-tuning. Although the proposed strategy exhibits suboptimal results on Potsdam (Table~\ref{tab:potsdam_tree_finetune_cbc}) when only RGB or IRRG is available, it still improves IoU by 0.2\% in the most challenging DSM-only case and by 0.4\% when all modalities are available, compared with the random modality dropout strategy. As shown in Fig.~\ref{fig:probabilities}, higher probabilities are assigned to scenarios with DSM modality, whereas scenarios including RGB and IRRG receive lower sampling emphasis. Therefore, the fine-tuning process is biased toward modality combinations that appear more disruptive in the pretrained latent space. As a result, the proposed strategy may provide clearer benefits in more challenging scenarios of the Potsdam image set, while yielding weaker gains in some relatively simpler configurations.

To show that the proposed training strategy is not limited to a specific architectural design, we also evaluate it using the state-of-the-art CMX model. The test set results are reported in Tables~\ref{tab:dstl_crop_finetune_cmx}, \ref{tab:potsdam_tree_finetune_cmx}, and \ref{tab:hunan_forest_finetune_cmx}. As shown in Table~\ref{tab:dstl_crop_finetune_cmx}, the proposed strategy outperforms other fine-tuning policies, achieving superior IoU scores on all scenarios. Notably, Tables~\ref{tab:potsdam_tree_finetune_cmx} and \ref{tab:hunan_forest_finetune_cmx} demonstrate large performance gains, indicating that the proposed training strategy generalizes particularly well to the CMX architecture under heterogeneous modality settings. In Table~\ref{tab:potsdam_tree_finetune_cmx}, when RGB is missing on Potsdam, the proposed strategy improves IoU and $F_1$ over the random modality dropout scheme by 2.7\% and 5.0\%, respectively. As listed in Table~\ref{tab:hunan_forest_finetune_cmx}, when only DEM is available on Hunan, our strategy achieves a noticeable improvement over LoRA-based adaptation. Specifically, it improves the IoU by 3.7\% and $F_1$ by 2.9\%. These results suggest that the proposed training strategy can generalize across different multimodal segmentation backbones. This may be because the strategy relies on scenario-induced distortions in the pretrained latent space, which provide a model-agnostic signal for guiding fine-tuning.

\begin{figure*}[t]
\centering
\includegraphics[width=\textwidth]{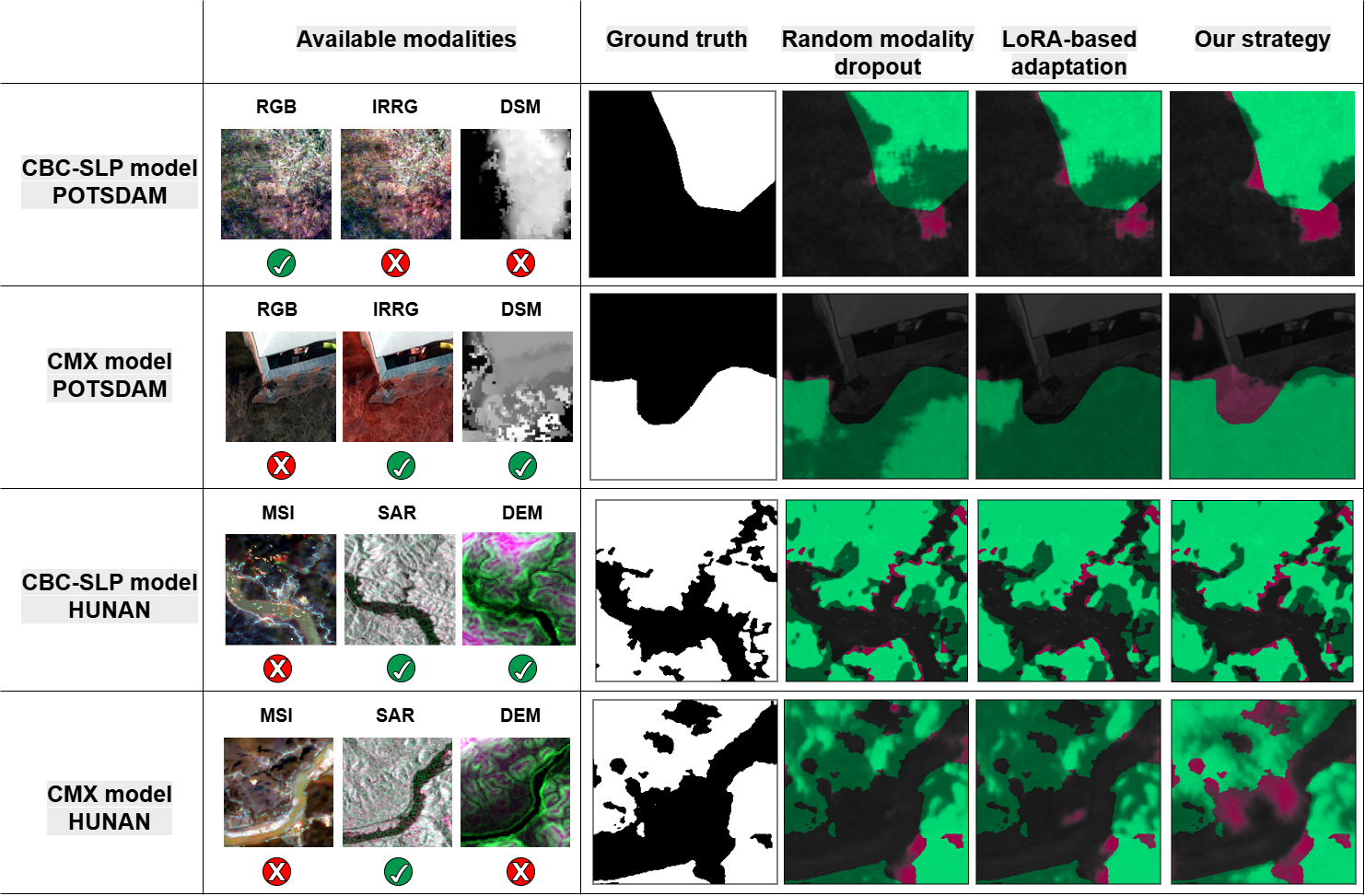}
\caption{Qualitative comparison of different fine-tuning settings under missing modality scenarios.}
\label{examples}
\end{figure*}

Fig.~\ref{examples} shows qualitative comparisons under different modality availability scenarios for the tree class on Potsdam and the forest class on Hunan. In these prediction maps, light green shows correctly predicted target pixels, dark green indicates missed pixels, and red marks falsely predicted regions. We evaluate the effectiveness of our training strategy in challenging scenarios, where important complementary information is unavailable.

In the first row, for CBC-SLP on Potsdam, only the RGB modality is available, while IRRG and DSM are missing. This makes the segmentation task more difficult, since infrared and elevation information are both crucial for accurate tree segmentation. Even under this scenario, our strategy better preserves the lower and right-side boundaries of the target region than the other fine-tuning strategies. In the second row, where RGB is missing, our strategy recovers a larger portion of the target, whereas the competing methods miss pixels around the lower central part of the region. In the third example, the absence of MSI removes rich spectral information, causing random modality dropout and LoRA-based adaptation strategies to miss many of the fine-grained details. In contrast, our strategy yields a more spatially consistent prediction by capturing complex boundary regions. For CMX on Hunan shown in the fourth row, only SAR is available, while MSI and DEM are both missing. This is the most difficult scenario among the shown examples, since both the spectral content and the topographic information are absent. Even in this challenging case, our strategy recovers more target pixels and yields smoother boundaries.

Taken together, these qualitative examples indicate that the proposed strategy yields more accurate boundaries, fewer missed target pixels, and better recovery of complex details under different modality availabilities. Since this trend is consistent for both CBC-SLP and CMX models on Potsdam and Hunan image sets, the proposed strategy appears to generalize across different multimodal segmentation backbones. A possible reason is that latent-space-guided scenario sampling encourages the model to focus more effectively on challenging modality configurations during fine-tuning.

\section{Conclusion}
We proposed a novel training strategy for multimodal semantic segmentation under missing modalities. We first measure the importance of each scenario with missing modalities according to the distortion it causes in the shared latent representation. We then use regularized kernel smoothing to convert these scenario statistics into a probability distribution. In this way, the proposed method replaces uniform random modality dropout with a training strategy that prioritizes more informative scenarios during fine-tuning. We further showed that this formulation is not limited to a specific architectural design, but can be applied across different multimodal segmentation backbones. This is demonstrated by the fact that the proposed strategy yields more consistent improvements than standard fine-tuning with random modality dropout and LoRA-based adaptation on the DSTL, Potsdam, and Hunan image sets, and across CBC-SLP, CBC, and CMX models. Finally, because the proposed strategy is driven by pretrained latent behavior, it may provide a framework for improving fine-tuning performance across different multimodal segmentation backbones when some modalities are unavailable.

\section*{Acknowledgment}

This study was supported by the Scientific and Technological Research Council of Turkey (TUBITAK) under the Grant Number 124E725. The authors thank TUBITAK for their support.

%\bibliographystyle{plain}

\bibliographystyle{cas-model2-names}


\bibliography{Mybib}

\vskip3pt

\bio{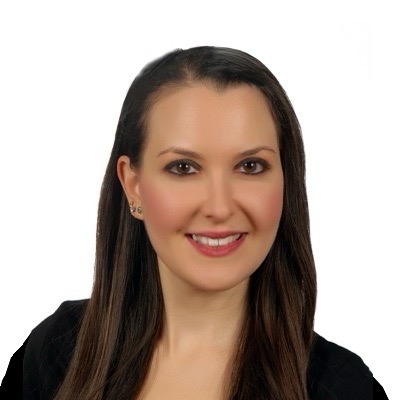}
Irem Ulku received B.Sc. degrees in both Electronics and Communication Engineering and in Industrial Engineering from Çankaya University (Ankara, Turkey) in 2009 and 2010, respectively, followed by an M.Sc. degree in Electrical and Electronics Engineering from the Middle East Technical University (Ankara, Turkey) in 2013. She then received her PhD degree in Electronics and Communication Engineering from Çankaya University in 2017. In 2019, she was a Research Associate at Imperial College London, United Kingdom. She is currently an Assistant Professor at the Department of Computer Engineering, Ankara University (Turkey). Her research interests include hyperspectral image processing and deep learning-based semantic segmentation.
\endbio

\bio{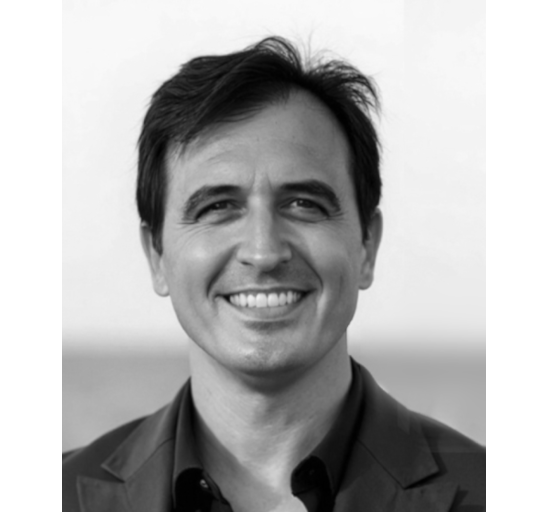}
Erdem Akagündüz received his B.Sc., M.Sc., and Ph.D. degrees in Electrical and Electronics Engineering from Middle East Technical University in 2001, 2004, and 2010, respectively. Between 2001–2008, he worked as a research assistant at METU Computer Vision and Intelligent Systems Laboratory; between 2008–2009, he was a research fellow at the University of York, UK; and from 2009–2016, he served as an embedded algorithm development engineer at ASELSAN Inc. In 2016, he worked as a postdoctoral researcher at the University of York again. After serving at Çankaya University between 2018–2021, he joined METU Informatics Institute, where he currently works. His research interests include deep learning applications and embedded deep learning. Dr. Akagündüz holds several international patents and publications in these areas.
\endbio

\bio{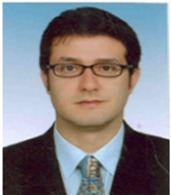}
Ömer Özgür Tanrıöver received the B.Sc. degree in Computer Engineering, the M.Sc. and Ph.D. degrees in Information Systems from Middle East Technical University, Ankara, Turkey. Earlier, he was as a Certified Information Systems Auditor (CISA) with the Information Management Department, Banking Regulation Agency of Turkey. He is currently an Associate Professor in the Computer Engineering Department of Ankara University. His current research interests include applications of deep  learning in medical informatics, remote sensing, human computer interaction, and information system security.
\endbio

\end{document}